%% file: iclr2027.tex
\documentclass{article}

\usepackage{iclr2027_conference,times}

\usepackage[utf8]{inputenc}
\usepackage[T1]{fontenc}
\usepackage{hyperref}
\usepackage{url}
\usepackage{booktabs}
\usepackage{amsmath}
\usepackage{amsfonts}
\usepackage{amssymb}
\usepackage{nicefrac}
\usepackage{microtype}
\usepackage{xcolor}
\usepackage{colortbl}
\usepackage{graphicx}
\usepackage{enumitem}
\usepackage{multirow}
\usepackage{listings}

\newcommand{\gravity}{\textsc{Gravity}}

\title{GRAVITY: Architecture-Agnostic Structured Anchoring for Long-Horizon Conversational Memory}

\author{%
Yushi Sun$^{1,*}$ \And
Bowen Cao$^{2,*}$ \And
Dong Fang$^{1,\dagger}$ \And
Lingfeng Su$^{1}$ \And
Wai Lam$^{2}$ \\
$^{1}$LIGHTSPEED, Shenzhen, China \\
$^{2}$The Chinese University of Hong Kong, Hong Kong, China \\
$^{*}$Equal contribution. Work done during Bowen Cao's internship at Tencent. \\
$^{\dagger}$Corresponding author: \texttt{df572@outlook.com}
}

\iclrfinalcopy

\hypersetup{
  colorlinks=true,
  linkcolor=black,
  citecolor=black,
  urlcolor=black,
  pdfauthor={Anonymous Authors},
  pdftitle={GRAVITY: Architecture-Agnostic Structured Anchoring for Long-Horizon Conversational Memory}
}

\begin{document}

\maketitle
\lhead{}

\begin{abstract}
Long-horizon memory systems increasingly improve how evidence is stored and retrieved, yet the generator must still reason over fragments whose cross-session relationships are implicit. We study generation-time memory organization as a distinct design dimension and introduce \gravity{} (\textbf{G}eneration-time \textbf{R}elational \textbf{A}nchoring \textbf{V}ia \textbf{I}njected \textbf{T}opological Memor\textbf{Y}), a host-independent auxiliary memory layer. \gravity{} consolidates raw dialogue into entity profiles, temporal event traces, and cross-session topic summaries, then retrieves and injects query-relevant records through the prompt interface. Across five heterogeneous memory systems on LongMemEval and LoCoMo, it improves every host--benchmark baseline under two distinct LLM configurations. Controlled analyses separate gains from organizing already available evidence and from consolidating information across the full history. Under a matched LightMem pipeline, the entity--event--topic representation reaches 83.9\% on LoCoMo, 3.6\% above the strongest of six alternative auxiliary representations. These results show that generation-time structure is a portable complement to existing memory retrieval, while its interaction with host evidence depends on the benchmark and host.
\end{abstract}

\input{sections/introduction}
\input{sections/related_work}
\input{sections/methodology}
\input{sections/experiments}
\input{sections/discussion}
\input{sections/conclusion}

\clearpage
\section*{AI use statement}
Generative AI models are experimental components of this work: they are used for anchor extraction, answer generation, and automated evaluation as documented in Sections~\ref{sec:method} and~\ref{sec:experiments} and in the appendix. Generative AI tools were also used to refine the language of the manuscript. The authors reviewed the manuscript changes and manually verified the reported results, claims, and references. The authors take responsibility for the final content of this work.

\section*{Ethics statement}
The experiments use existing synthetic or curated conversational-memory benchmarks and do not collect new human-subject data. Persistent memory can nevertheless expose sensitive information or preserve extraction errors. Deployments should obtain user consent, enforce access control, and support inspection, correction, and deletion of stored records. Appendix~\ref{app:impacts} discusses these risks and mitigations.

\section*{Reproducibility statement}
The method and inference pipeline are specified in Section~\ref{sec:method}. Section~\ref{sec:experiments} reports the evaluation protocol, benchmarks, hosts, metrics, and efficiency measurements. The appendix provides detailed settings, ablations, prompts, and additional analyses needed to reproduce the reported results.

\bibliography{reference}
\bibliographystyle{iclr2027_conference}

\clearpage
\appendix
\input{sections/appendix}

\end{document}

%% file: sections/introduction.tex
\section{Introduction}
\label{sec:intro}

Long-horizon conversational agents must recover facts, changes, and dependencies spread across many sessions. Existing memory systems address this challenge with increasingly capable storage and retrieval mechanisms, including vector stores, linked notes, entity graphs, temporal graphs, and hierarchical compression~\citep{packer2023memgpt,park2023generative,chhikara2025mem0,rasmussen2025zep,xu2025amem,fang2026lightmem}. These systems differ substantially in what they store and how they search, but they share a final step: the generator receives a bounded context and must reconstruct the relationships among the returned items.

This final step creates a separate bottleneck. A retrieved fragment may mention an entity without its attributes, an event without the earlier event that explains it, or one session from a topic that develops over months. Better retrieval can expose more relevant fragments, but it does not by itself organize their relational, temporal, and thematic connections. The distinction is measurable. In our oracle diagnostic, the prompt already contains all annotated evidence, yet accuracy drops from 84.9\% with evidence alone to 75.6\% when that evidence is scattered among distractors. In a matched control, structuring only the sessions already retrieved by a host improves accuracy from 80.9\% to 82.4\%, while constructing the same types of records over the full history reaches 85.5\%. Long-horizon memory therefore involves both \emph{evidence access} and \emph{evidence organization}.

We introduce \gravity{} (\textbf{G}eneration-time \textbf{R}elational \textbf{A}nchoring \textbf{V}ia \textbf{I}njected \textbf{T}opological Memor\textbf{Y}), a host-independent auxiliary memory layer designed around this distinction. It consolidates raw dialogue into three recurrent structures: \emph{entity profiles} connect attributes and relationships across mentions, \emph{event traces} preserve temporal order and outcomes, and \emph{topic summaries} capture cross-session arcs. At inference, \gravity{} retrieves query-relevant records from each bank and injects them beside the host's own memories. Construction depends only on raw dialogue, and integration uses the prompt and query interfaces, so the same anchor store can augment heterogeneous hosts without changing their internal storage or retrieval algorithms. The contribution is the typed, cross-session consolidation and selection pipeline, not the surface labels used to display its records.

Our evaluation follows the resulting evidence chain. Under the primary configuration, \gravity{} improves all five memory systems on both LongMemEval and LoCoMo, with average gains of 9.2\% and 7.5\%. Changing both the generation and evaluation models preserves positive gains in all ten host--benchmark pairs. Applying two additional evaluators to the same primary outputs also preserves all 30 host--benchmark--evaluator comparisons. A controlled representation study further shows that entity--event--topic anchoring reaches 83.9\% on LoCoMo, compared with 80.3\% for the strongest of six matched alternatives and 78.0\% for a free-form summary. Anchor-only and error analyses qualify the result: the external store is strong by itself, but host evidence adds consistently beyond it only on LongMemEval, and extraction errors can introduce over-summarization or temporal noise.

\noindent\textbf{Contributions.}
\begin{enumerate}[leftmargin=1.2em,itemsep=0pt,topsep=2pt,parsep=0pt]
    \item We formulate generation-time memory organization as a bottleneck distinct from retrieval and quantify both organization and full-history consolidation effects with oracle, same-evidence, and evidence-coverage controls.
    \item We propose a host-independent entity--event--topic memory layer that consolidates cross-session structure and attaches to existing systems through query expansion and prompt-level context injection.
    \item We establish transfer across five hosts, two benchmarks, two model configurations, and three evaluators, and compare the proposed representation against six matched structured alternatives while reporting where host and anchor evidence reinforce or interfere with one another.
\end{enumerate}

%% file: sections/related_work.tex
\section{Related Work}
\label{sec:related}

\noindent\textbf{Conversational memory systems.}
Long-horizon agents use explicit memory to preserve information beyond the current context~\citep{park2023generative,packer2023memgpt}. Recent systems span atomic linked notes (A-Mem)~\citep{xu2025amem}, vector and entity-graph storage (Mem0)~\citep{chhikara2025mem0}, temporal knowledge graphs (ZEP)~\citep{rasmussen2025zep}, hierarchical cognitive graphs (LiCoMemory)~\citep{huang2025licomemory}, and hierarchical compression with consolidation (LightMem)~\citep{fang2026lightmem}. These systems jointly define the host set in our evaluation. Their internal representations can be highly structured, but their retrieved units and generation contexts remain specific to each architecture. \gravity{} constructs one external representation from raw dialogue and uses the same prompt-level interface across all five hosts.

\noindent\textbf{Structured retrieval and auxiliary representations.}
GraphRAG builds entity communities and community summaries~\citep{edge2024graphrag}; RAPTOR retrieves from a hierarchy of recursive summaries~\citep{sarthi2024raptor}; HippoRAG2 combines passages and knowledge-graph triples through personalized PageRank~\citep{gutierrez2025hipporag2}; LightRAG uses dual-level graph retrieval~\citep{guo2024lightrag}; and Generative Semantic Workspaces maintain evolving actor-centric state~\citep{holur2024gsw}. These works provide distinct structured representations that can be adapted as auxiliary generation context. We compare adaptations of these representation schemes within one LightMem pipeline, holding the host, model configuration, prompt position, and retrieval budget fixed. This controlled comparison isolates auxiliary representation choice; it is not an end-to-end reproduction of the original systems.

\noindent\textbf{Prompt-level augmentation.}
Retrieval-augmented generation conditions model outputs on evidence retrieved from external knowledge sources~\citep{lewis2020rag,gao2024rag_survey}, while adapters and tools extend models through lightweight modules or external calls~\citep{houlsby2019adapters,schick2024toolformer}. \gravity{} follows the non-intrusive interface of retrieval augmentation but focuses on a different variable: cross-session consolidation of relational, temporal, and thematic content before generation. This separates the external memory representation from the host's retrieval architecture.

%% file: sections/methodology.tex
\section{Method}
\label{sec:method}

\begin{figure}[t]
\centering
\includegraphics[width=\linewidth]{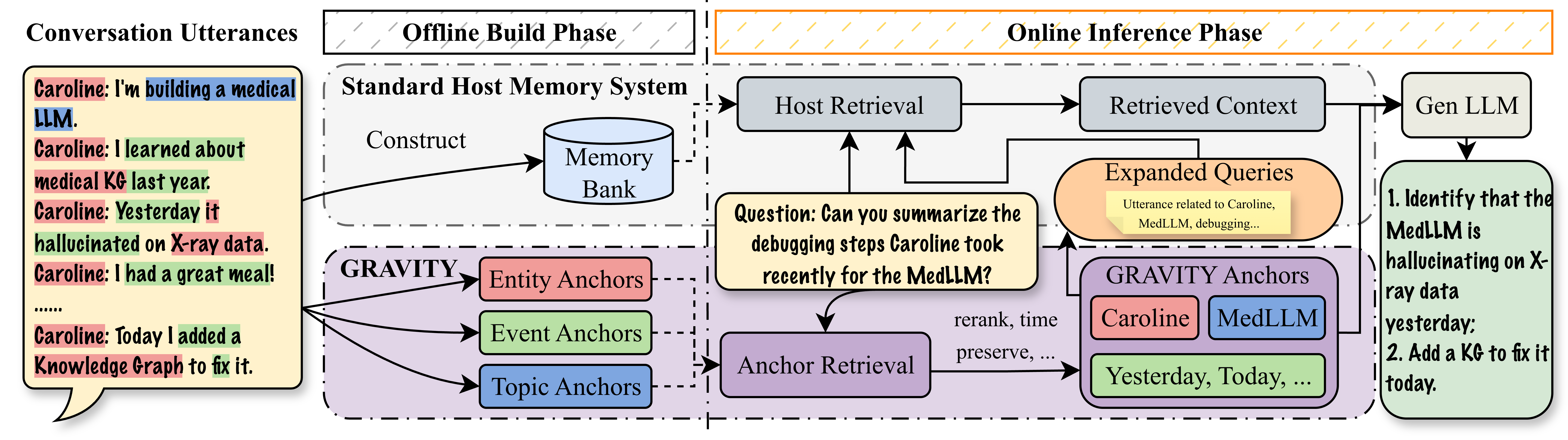}
\caption{\gravity{} builds entity, event, and topic banks from raw dialogue independently of the host. At inference, it retrieves query-relevant anchors, expands the host query, and injects the selected records beside the host memories.}
\label{fig:overview}
\end{figure}

\gravity{} is an external structured memory built directly from raw dialogue. It can supply generation context alone, but its intended use is to complement a host memory system: the host retains its original storage and retrieval pipeline, while \gravity{} contributes cross-session records selected for the current query. Figure~\ref{fig:overview} shows the two phases.

\subsection{Design Principle}
\label{sec:framework}

A long conversation contains several views of the same history. \emph{Relational structure} links entities to attributes and to one another. \emph{Temporal structure} connects events through order, duration, and state change. \emph{Thematic structure} groups episodes into topics that recur across sessions. A bounded list of retrieved fragments may contain the relevant statements without making these connections explicit. We therefore materialize the three views before query time rather than asking the generator to reconstruct them for every question.

Let $\mathcal{M}(q)$ denote the host memories retrieved for query $q$. \gravity{} maintains entity, event, and topic banks $\mathcal{A}_E$, $\mathcal{A}_V$, and $\mathcal{A}_T$, and selects a query-specific anchor set
\begin{equation}
\mathcal{S}(q)=\mathcal{A}_E(q)\cup\mathcal{A}_V(q)\cup\mathcal{A}_T(q).
\end{equation}
The generator receives $\mathcal{C}(q)=\mathcal{M}(q)\cup\mathcal{S}(q)$. The method changes the content available at generation time, not the host's internal memory architecture. Typed records and cross-session consolidation define the structure; the visible block labels in the prompt are only a serialization choice.

\paragraph{Why these three views?}
The banks factor three kinds of dependency that are repeatedly required by long-horizon questions. Entity records answer which attributes and relations belong to the same persistent object; event records answer what changed, in what order, and with what outcome; topic records answer which separated episodes form one continuing narrative. The factorization is deliberately overlapping. An employment change, for example, updates a person's entity state, appears as an event with a date and outcome, and belongs to a longer career topic. The overlap lets retrieval recover the same evidence through different query cues, while typed fields prevent the entire history from being compressed into one undifferentiated summary. This design is evaluated rather than assumed: the component ablation tests each view and their combinations, and the matched representation study compares the tri-view store with free-form summaries, hierarchical summaries, graph triples, and other structured alternatives.

\paragraph{What is shared across hosts?}
A host adapter exposes only two operations: retrieve a fixed number of memories for a text query and serialize those memories for generation. \gravity{} neither reads nor rewrites the host's internal graph, embeddings, clustering, or update policy. Its own JSON records, embeddings, thresholds, and retrieval outputs are identical across hosts for a given conversation and query. Host-specific differences arise only after expanded queries are passed through the host's public retrieval path and the returned host memories are combined with the selected anchors. This boundary makes ``architecture-agnostic'' an interface property that can be tested directly, rather than a claim that all underlying memory architectures are equivalent.

\subsection{Building the Anchor Banks}
\label{sec:build}

Given utterances $\mathcal{U}=\{u_1,\ldots,u_N\}$ with speaker, session, and timestamp metadata, three batched extractors update persistent records. Each record retains source provenance for auditing, but inference uses the consolidated record rather than fetching its source utterances. The prompts and full field specifications appear in Appendix~\ref{app:prompts}.

\paragraph{Entity anchors.}
\label{sec:entity}
The entity bank maintains a dynamic profile for each person, organization, location, project, or other persistent object. A profile contains key--value attributes, typed relations, status changes, a timeline, and source identifiers. Batch-level extractions are merged with confidence-weighted updates. Consistent values accumulate support; when values conflict, the better-supported value becomes current and the previous value moves to a historical log. Relations are stored in both directions and deduplicated during consolidation. Recurrent co-occurrences provide candidates for latent links, and a compact profile summary becomes the retrieval representation. This process combines mentions without discarding the order in which an entity's state changes.

\paragraph{Event anchors.}
\label{sec:event}
The event bank represents each event as a 4W1O tuple, \texttt{(Who, What, When, Where, Outcome)}, together with an event type and importance. The \texttt{When} field separates absolute time, relative expressions, duration, and recurrence. Events sharing participants or topics are linked into chronological traces. For duplicate detection, we average participant overlap, content-word overlap, normalized-time equality, and location equality, and merge pairs above a fixed threshold. The merged record keeps the more complete fields and all available provenance. Compared with an entity timeline, an event trace preserves interactions among several entities and the outcomes linking one event to the next.

\paragraph{Topic anchors.}
\label{sec:topic}
The topic bank groups semantically related utterances across sessions. The extractor assigns each utterance a topic label and keywords in overlapping batches, and clusters are merged by keyword and utterance overlap. A second pass produces a structured summary with the topic label, participants, temporal span, narrative synopsis, key facts, and keywords. Separating assignment from summarization prevents a locally coherent batch summary from fixing the global topic boundary too early. If a conversation contains no usable records for a bank, inference simply skips that bank.

These updates are incremental: new batches modify existing profiles, append or merge events, and extend topic clusters without rebuilding the host memory. The experiments use static benchmark histories so one completed anchor store can be reused across hosts; the same update operators support prefix-by-prefix construction.

\paragraph{Provenance and conflicting records.}
Every extracted field is paired with source identifiers before consolidation. Agreement across batches raises support for a value; disagreement does not erase the older observation but moves it into a time-indexed history. This distinction matters for questions about updates: ``current employer'' should resolve to the latest supported value, whereas ``previous employer'' requires retaining the superseded one. Event merging is more conservative because two semantically similar descriptions may refer to repeated activities rather than duplicates; temporal and participant agreement are therefore required in addition to content overlap. Provenance is retained in the stored artifacts and used for evidence-coverage analysis, although the current generator receives the consolidated fields rather than explicit source quotations. The latter choice keeps prompts compact but creates the abstraction failures studied in Section~\ref{sec:error}.

\subsection{Retrieval and Prompt-Level Integration}
\label{sec:inference}

For each query, every bank first retrieves candidates with its native keys, then reranks them by cosine similarity $s(q,a)=\cos(\mathbf{e}_q,\mathbf{e}_a)$ between query and record embeddings. We retain the top $K$ records per bank above a similarity threshold. Entity candidates use name and attribute matches, event candidates use participants and keywords, and topic candidates use labels and keywords. For temporal queries, reserved event slots preserve records whose normalized time matches the query even when lexical similarity is low. This prevents a phrase such as ``recently'' from excluding a lexically distant but correctly dated event.

The selected records support two operations. First, each bank forms expanded retrieval queries from its fields: entity names are combined with attributes and relations, events with participants and temporal references, and topics with key facts and participants. A round-robin merge draws from the three banks until nine queries are selected. The host executes these queries through its existing search interface, and newly retrieved items replace the lowest-similarity items so the host retrieval count remains fixed. Second, the selected anchors are serialized beside the host memories as topic summaries, entity profiles, and event records. The generator is instructed to treat source memories as primary evidence and the anchors as consolidated context for linking, disambiguation, and gap filling. Appendix~\ref{app:settings} specifies the exact retrieval budgets and injected fields.

\label{sec:integration}
Host independence follows from two boundaries: construction reads only raw dialogue, and integration uses only query and prompt interfaces. The anchor files contain no host-specific embedding, segmentation, graph, or storage object. The same store is therefore built once and attached unchanged to all evaluated hosts.

%% file: sections/experiments.tex
\section{Experiments}
\label{sec:experiments}

\subsection{Setup}
\label{sec:setup}

\noindent\textbf{Benchmarks and metric.}
We evaluate on LoCoMo~\citep{maharana2024evaluating}, using its 1{,}540 non-adversarial multi-session questions, and LongMemEval~\citep{wu2025longmemeval}, using all 500 questions including 30 abstention cases. We report LLM-judge accuracy; category definitions and full breakdowns are in Appendix~\ref{app:datasets}.

\noindent\textbf{Hosts.}
We attach the same anchor store to five systems spanning distinct memory designs: LightMem~\citep{fang2026lightmem}, A-Mem~\citep{xu2025amem}, Mem0~\citep{chhikara2025mem0}, LiCoMemory~\citep{huang2025licomemory}, and ZEP~\citep{rasmussen2025zep}. The hosts cover hierarchical compression, linked atomic notes, vector memory with graph augmentation, hierarchical cognitive graphs, and temporal knowledge graphs.

\noindent\textbf{Model configurations.}
The primary controlled comparison uses the frozen GPT-4o-mini configuration of the original host-aligned study for memory construction, answer generation, anchor extraction, and evaluation. We also evaluate the complete five-host, two-benchmark matrix under a second configuration, using Claude Sonnet 4.6 for generation and anchor extraction and GPT-5.6 sol as a cross-family judge. Host settings, retrieval budgets, prompts, and fairness controls are given in Appendix~\ref{app:settings}.

\subsection{Effectiveness and Transfer}

\begin{table}[t]
\centering
\caption{Primary GPT-4o-mini results. LLM-judge accuracy (\%) before and after adding \gravity{}. LongMemEval Overall covers all 500 questions.}
\label{tab:main}
\footnotesize
\setlength{\tabcolsep}{6pt}
\renewcommand{\arraystretch}{1.05}
\begin{tabular}{@{}l ccc ccc@{}}
\toprule
& \multicolumn{3}{c}{\textbf{LongMemEval}} & \multicolumn{3}{c}{\textbf{LoCoMo}} \\
\cmidrule(lr){2-4} \cmidrule(lr){5-7}
\textbf{System} & Base & +G & $\Delta$ & Base & +G & $\Delta$ \\
\midrule
ZEP~\citep{rasmussen2025zep} & 48.2 & 60.4 & \textbf{+12.2} & 54.7 & 61.8 & +7.1 \\
Mem0~\citep{chhikara2025mem0} & 52.6 & 64.0 & +11.4 & 51.0 & 59.1 & +8.1 \\
A-Mem~\citep{xu2025amem} & 53.8 & 63.2 & +9.4 & 65.3 & 70.9 & +5.6 \\
LiCoMemory~\citep{huang2025licomemory} & 57.4 & 66.6 & +9.2 & 55.7 & 66.2 & \textbf{+10.5} \\
LightMem~\citep{fang2026lightmem} & \textbf{68.8} & \textbf{72.6} & +3.8 & \textbf{70.1} & \textbf{75.8} & +5.7 \\
\midrule
\rowcolor{gray!8}
\textit{Average} & 56.2 & 65.4 & +9.2 & 59.3 & 66.8 & +7.5 \\
\bottomrule
\end{tabular}
\end{table}

Table~\ref{tab:main} shows positive gains for every host on both benchmarks. Average accuracy rises by 9.2\% on LongMemEval and 7.5\% on LoCoMo. The improvement spans systems whose internal memories are already structured, including Mem0 and ZEP, as well as compression and note-based systems. This supports portability of the external construction and integration interface rather than a claim that all host memories interact with anchors identically.

The cross-model evaluation in Table~\ref{tab:cross_model_main} also improves all ten host--benchmark pairs, with average gains of 15.2\% on LongMemEval and 18.1\% on LoCoMo. Applying Claude Sonnet 4.6 and GPT-5.6 sol as alternative evaluators to the same primary outputs preserves a positive gain in all 30 host--benchmark--evaluator comparisons. The effect therefore transfers across both the generator/extractor configuration and the evaluator. Detailed task results are in Appendix~\ref{app:main_results}.

The category breakdown clarifies that the gain is not confined to one question type. Under the primary configuration, all five hosts improve on every non-adversarial LoCoMo category, including multi-hop and temporal questions. LongMemEval is more heterogeneous: the largest changes occur in preference, multi-session, and temporal tasks for hosts that initially miss those capabilities, while assistant-memory, knowledge-update, and abstention results can move in either direction. For example, LightMem's overall LongMemEval accuracy increases by 3.8\% even though its abstention subset decreases from 60.0\% to 53.3\%. We therefore treat aggregate transfer across hosts and benchmarks as the central result, and retain the task-level tables in Appendix~\ref{app:main_results} to expose these local regressions.

\begin{table}[t]
\centering
\caption{Cross-model evaluation with Claude Sonnet 4.6 for generation and anchor extraction and GPT-5.6 sol for evaluation (accuracy, \%).}
\label{tab:cross_model_main}
\footnotesize
\setlength{\tabcolsep}{5pt}
\begin{tabular}{@{}l ccc ccc@{}}
\toprule
& \multicolumn{3}{c}{\textbf{LongMemEval}} & \multicolumn{3}{c}{\textbf{LoCoMo}} \\
\cmidrule(lr){2-4} \cmidrule(lr){5-7}
\textbf{System} & Base & +G & $\Delta$ & Base & +G & $\Delta$ \\
\midrule
ZEP & 57.6 & 85.8 & +28.2 & 56.6 & 79.8 & +23.2 \\
Mem0 & 66.8 & 81.0 & +14.2 & 53.4 & 79.3 & +25.9 \\
A-Mem & 82.4 & 87.0 & +4.6 & 71.2 & 83.8 & +12.6 \\
LiCoMemory & 72.4 & \textbf{88.4} & +16.0 & 58.1 & 80.7 & +22.6 \\
LightMem & 70.4 & 83.4 & +13.0 & 77.8 & \textbf{83.9} & +6.1 \\
\midrule
\rowcolor{gray!8}\textit{Average} & 69.9 & 85.1 & +15.2 & 63.4 & 81.5 & +18.1 \\
\bottomrule
\end{tabular}
\end{table}

\subsection{Component and Representation Controls}
\label{sec:ablation}

\begin{table}[t]
\centering
\caption{Primary LightMem ablation (accuracy, \%). E/V/T denote Entity, Event, and Topic; Sum is a free-form summary; $-$exp and $-$rrk remove query expansion and reranking.}
\label{tab:ablation}
\scriptsize
\setlength{\tabcolsep}{2.5pt}
\renewcommand{\arraystretch}{1.05}
\begin{tabular}{@{}l c c ccc ccc c cc@{}}
\toprule
& & & \multicolumn{3}{c}{\textit{Single}} & \multicolumn{3}{c}{\textit{Pairwise}} & & \multicolumn{2}{c}{\textit{Design}} \\
\cmidrule(lr){4-6} \cmidrule(lr){7-9} \cmidrule(lr){11-12}
\textbf{Benchmark} & \textbf{Base} & \textbf{+Sum} & \textbf{+E} & \textbf{+V} & \textbf{+T} & \textbf{+EV} & \textbf{+ET} & \textbf{+VT} & \cellcolor{gray!8}\textbf{+EVT} & \textbf{$-$exp} & \textbf{$-$rrk} \\
\midrule
LoCoMo & 70.1 & 71.4 & 72.3 & 70.8 & 72.8 & 72.1 & 74.0 & 74.0 & \cellcolor{gray!8}\textbf{75.8} & 75.3 & 72.4 \\
LME & 68.8 & 69.0 & 71.6 & 70.6 & 70.0 & 70.0 & 71.0 & 69.0 & \cellcolor{gray!8}\textbf{72.6} & \textbf{73.0} & 68.8 \\
\bottomrule
\end{tabular}
\end{table}

The free-form summary adds only 1.3\% on LoCoMo and 0.2\% on LongMemEval, whereas the full tri-view configuration adds 5.7\% and 3.8\%. No single or pairwise view matches the full configuration on either benchmark. Removing query expansion has little effect and slightly improves LongMemEval, while removing reranking erases the LongMemEval gain. The main mechanism is therefore the construction and selection of typed cross-session records; expanded host retrieval is secondary. The cross-model evaluation in Table~\ref{tab:cross_model_ablation} preserves the advantage of the full tri-view configuration.

\begin{table}[t]
\centering
\caption{Cross-model ablation with LightMem (accuracy, \%).}
\label{tab:cross_model_ablation}
\footnotesize
\setlength{\tabcolsep}{5pt}
\begin{tabular}{@{}l cc@{}}
\toprule
\textbf{Configuration} & \textbf{LoCoMo} & \textbf{LongMemEval Overall} \\
\midrule
Base & 77.8 & 70.4 \\
+ Summary & 78.0 & 72.0 \\
+ Entity & 78.4 & 77.6 \\
+ Event & 79.9 & 78.6 \\
+ Topic & 81.4 & 77.8 \\
+ Entity + Event & 80.8 & 79.0 \\
+ Entity + Topic & 82.2 & 82.0 \\
+ Event + Topic & 81.9 & 81.6 \\
\rowcolor{gray!8}
+ All three & \textbf{83.9} & \textbf{83.4} \\
$-$ query expansion & 83.2 & \textbf{83.4} \\
$-$ reranking & 83.6 & 82.2 \\
\bottomrule
\end{tabular}
\end{table}

\begin{table}[t]
\centering
\caption{Controlled LoCoMo diagnostics with LightMem under the cross-model configuration. Representation rows use a fixed pipeline; source rows separate organization from evidence access.}
\label{tab:controlled}
\footnotesize
\setlength{\tabcolsep}{8pt}
\begin{tabular}{@{}l cc@{}}
\toprule
\multicolumn{3}{@{}l}{\textit{(a) Auxiliary representation in a fixed pipeline}} \\
\textbf{Representation} & \textbf{Accuracy} & \textbf{$\Delta$} \\
\midrule
None (LightMem) & 77.8 & -- \\
Free-form summary & 78.0 & +0.2 \\
GraphRAG-style~\citep{edge2024graphrag} & 77.2 & $-0.6$ \\
RAPTOR-style~\citep{sarthi2024raptor} & 79.1 & +1.3 \\
HippoRAG2-style~\citep{gutierrez2025hipporag2} & 79.1 & +1.3 \\
LightRAG-style~\citep{guo2024lightrag} & 80.1 & +2.3 \\
GSW-style~\citep{holur2024gsw} & 80.3 & +2.5 \\
\rowcolor{gray!8}Entity--event--topic & \textbf{83.9} & \textbf{+6.1} \\
\midrule
\multicolumn{3}{@{}l}{\textit{(b) Evidence source and organization}} \\
\textbf{Generation context} & \textbf{Accuracy} & \textbf{$\Delta$} \\
\midrule
Host retrieval only & 80.9 & -- \\
Records from retrieved sessions only & 77.1 & $-3.8$ \\
Host + records from retrieved sessions & 82.4 & +1.5 \\
Records from full history only & 80.3 & $-0.6$ \\
Host + records from full history & \textbf{85.5} & \textbf{+4.6} \\
\bottomrule
\end{tabular}
\end{table}

Panel (a) of Table~\ref{tab:controlled} compares six alternative auxiliary representations after adapting each to the same host, model, prompt position, and retrieval budget. Entity--event--topic anchoring exceeds the strongest alternative by 3.6\%. These rows test representation choice rather than full end-to-end implementations of the cited systems.

Panel (b) separates two mechanisms. Building records only from sessions already retrieved by the host adds 1.5\%, even though it introduces no source evidence outside the host prompt. This is the organization effect. Constructing the records over the full history adds another 3.1\%, which reflects broader evidence access. Removing the raw host entries from either condition lowers accuracy by 5.2--5.3\%: consolidated records help connect evidence but do not preserve every exact detail. A retrieval-side check agrees with this decomposition. Anchor provenance raises mean gold-session recall from 0.945 to 0.977 and the fraction of questions with complete gold-session coverage from 0.900 to 0.960. In the oracle diagnostic, where annotated evidence is fixed and only its arrangement changes, anchors recover 2.9\% when that evidence is scattered among distractors (Appendix~\ref{sec:appendix_oracle}).

\subsection{Memory-Source Interaction}
\label{sec:interaction}
\label{sec:oracle}

\begin{table}[t]
\centering
\caption{Full-benchmark source diagnostic under the cross-model configuration (accuracy, \%). $H$: host only; $A$: the shared anchor store only. The last column measures the host's contribution beyond anchors.}
\label{tab:anchor_only}
\scriptsize
\setlength{\tabcolsep}{4.5pt}
\renewcommand{\arraystretch}{1.02}
\begin{tabular}{@{}ll ccc rr@{}}
\toprule
\textbf{Benchmark} & \textbf{Host} & \textbf{$H$} & \textbf{$A$} & \textbf{$H{+}A$} & \textbf{$H{+}A-H$} & \textbf{$H{+}A-A$} \\
\midrule
\multirow{6}{*}{LoCoMo}
& ZEP          & 56.6 & 80.9 & 79.8 & +23.2 & $-1.1$ \\
& Mem0         & 53.4 & 80.9 & 79.3 & +25.9 & $-1.6$ \\
& A-Mem        & 71.2 & 80.9 & 83.8 & +12.6 & +2.9 \\
& LiCoMemory   & 58.1 & 80.9 & 80.7 & +22.6 & $-0.2$ \\
& LightMem     & 77.8 & 80.9 & 83.9 &  +6.1 & +3.0 \\
& \textit{Average} & 63.4 & 80.9 & 81.5 & +18.1 & +0.6 \\
\midrule
\multirow{6}{*}{LongMemEval}
& ZEP          & 57.6 & 79.0 & 85.8 & +28.2 & +6.8 \\
& Mem0         & 66.8 & 79.0 & 81.0 & +14.2 & +2.0 \\
& A-Mem        & 82.4 & 79.0 & 87.0 &  +4.6 & +8.0 \\
& LiCoMemory   & 72.4 & 79.0 & 88.4 & +16.0 & +9.4 \\
& LightMem     & 70.4 & 79.0 & 83.4 & +13.0 & +4.4 \\
& \textit{Average} & 69.9 & 79.0 & 85.1 & +15.2 & +6.1 \\
\bottomrule
\end{tabular}
\end{table}

Table~\ref{tab:anchor_only} shows two different comparisons that should not be conflated. Relative to each host baseline, $H{+}A$ improves all ten host--benchmark pairs by 4.6--28.2\%, supporting portability of the augmentation. Relative to anchors alone, host evidence adds 2.0--9.4\% for every LongMemEval host but has mixed value on LoCoMo. The shared store therefore functions as a strong standalone memory, but combining it with a host is not universally superadditive. The benchmark difference is consistent with LongMemEval's exact updates and abstention cases benefiting from raw retrieved evidence, while LoCoMo often rewards the consolidated cross-session view directly.

\subsection{Evaluator Robustness and Practical Cost}
\label{sec:efficiency}

\begin{table}[t]
\centering
\caption{Evaluator sensitivity and build trade-offs. Panel (a) applies three evaluators to identical outputs from the primary runs and reports the mean anchoring gain. Panel (b) reports mean LoCoMo accuracy across five hosts; build cost is per conversation.}
\label{tab:robust_cost}
\footnotesize
\setlength{\tabcolsep}{7pt}
\begin{tabular}{@{}l ccc@{}}
\toprule
\multicolumn{4}{@{}l}{\textit{(a) Evaluator sensitivity: average $H{+}A-H$ (\%)}} \\
\textbf{Benchmark} & \textbf{GPT-4o-mini} & \textbf{Sonnet 4.6} & \textbf{GPT-5.6 sol} \\
\midrule
LongMemEval & +9.2 & +9.5 & +8.3 \\
LoCoMo & +7.5 & +6.4 & +6.2 \\
\midrule
\multicolumn{4}{@{}l}{\textit{(b) Anchor construction variants on LoCoMo}} \\
\textbf{Variant} & \textbf{Build tokens (K)} & \textbf{Build time (s)} & \textbf{Mean accuracy} \\
\midrule
No anchors & -- & -- & 59.3 \\
Default separate calls & 193 & 557 & \textbf{66.8} \\
Parallel separate calls & 193 & \textbf{280} & \textbf{66.8} \\
Fused triple extraction & \textbf{48} & 453 & 64.1 \\
Qwen-3-8B extraction & 286 & 667 & 65.6 \\
\bottomrule
\end{tabular}
\end{table}

Applying three evaluators to identical outputs yields a positive delta in all 30 host--benchmark--evaluator comparisons, with similar average magnitudes across evaluators (Table~\ref{tab:robust_cost}(a)). The default anchor store is built once and reused across hosts. The construction variants expose three distinct deployment trade-offs. Parallel module calls preserve the default outputs and token count while reducing wall-clock time from 557 to 280 seconds. Fusing the three extractors reduces build tokens by 75\%, from 193K to 48K, at a 2.7\% reduction in mean anchored accuracy. Replacing only anchor extraction with Qwen-3-8B removes API dependence and reaches 65.6\%, compared with 66.8\% for the default build. At inference, anchors add less than one second for most hosts; graph-based LiCoMemory and ZEP incur larger query-expansion overhead. Appendix~\ref{app:efficiency} reports per-host construction and inference costs, parameter sensitivity, and complete variant results.

\subsection{Failure Analysis}
\label{sec:error}

\begin{table}[t]
\centering
\caption{Manual taxonomy of 50 sampled losses under the primary configuration.}
\label{tab:error_taxonomy}
\footnotesize
\setlength{\tabcolsep}{7pt}
\begin{tabular}{@{}l c l@{}}
\toprule
\textbf{Failure type} & \textbf{Share} & \textbf{Observed effect} \\
\midrule
Over-summarization & 38\% & Specific facts collapse into broader descriptions \\
Temporal-slot error & 32\% & Relative or ambiguous time is normalized incorrectly \\
Entity confusion & 20\% & An attribute is merged into the wrong profile \\
Topic over-generalization & 10\% & A distinctive event is replaced by a broad theme \\
\bottomrule
\end{tabular}
\end{table}

Question-level analysis finds positive net gains for all hosts, with net improvements of 85--160 questions and an overall gain-to-loss ratio of 2.2:1. One representative LightMem case asks for Nate's favorite dish from a cooking show. The baseline answers ``not specified,'' whereas anchoring recovers ``coconut milk ice cream.'' The relevant attribute appears only in passing within a longer exchange: host retrieval surfaces the surrounding show discussion but misses the exact turn, while the entity profile consolidates the scattered mentions into one record. Table~\ref{tab:error_taxonomy} shows the corresponding cost of abstractive consolidation. A representative loss occurs when an ambiguous temporal expression is normalized to a specific but incorrect date and the generator trusts the record over a safer host answer. The effect is also host-specific: pairwise Jaccard similarity is only 0.09--0.17 for gain sets and 0.04--0.13 for loss sets, so the shared store repairs different blind spots in different hosts. These results motivate provenance-aware verification, conflict detection, and stronger temporal normalization rather than more aggressive compression. Appendix~\ref{app:error_analysis} provides the full counts, cases, and overlap matrices.

%% file: sections/discussion.tex
\section{Discussion and Limitations}
\label{sec:discussion}

\paragraph{What architecture-agnosticism does and does not mean.}
\gravity{} is architecture-agnostic at the interface level: one store is constructed without host internals and improves every evaluated host baseline. This does not imply identical host--anchor interactions. LongMemEval benefits from combining raw host evidence with anchors, whereas three LoCoMo hosts fall slightly below anchors alone. Likewise, removing prompt block labels does not consistently hurt accuracy. The supported design principle is typed cross-session consolidation and query-specific selection, not a universal advantage from any second memory or from visibly structured formatting. The full source diagnostic is important here: portability is established by positive $H{+}A-H$ across all ten settings, while complementarity is a separate property measured by $H{+}A-A$.

\paragraph{Abstraction and verification.}
The main limitation is that anchors are abstractive records produced by an LLM. Consolidation improves coverage and organization but can discard exact wording, normalize time incorrectly, or merge attributes across entities. The current prompt tells the generator to treat host memories as primary evidence, yet a concise anchor can still dominate a longer and more accurate source passage. A natural extension is retrieval-time verification: attach each selected field to its provenance, retrieve the supporting utterance on demand, and suppress or qualify an anchor when the source and consolidated value conflict. Our replacement controls indicate why this matters. Records alone underperform the corresponding records plus raw host evidence by about five points, so removing source memories in favor of a purely compressed store would trade away exact-detail reliability.

\paragraph{Evaluation scope.}
The two benchmarks cover multi-session factual, temporal, preference, update, and abstention questions, but they are synthetic or curated QA environments rather than deployed assistants. They do not measure privacy deletion, user correction, adversarial memory injection, or the long-term accumulation of extraction errors. The primary evaluation uses an LLM evaluator; applying two additional model families to the same outputs preserves every gain direction. Human assessment would further resolve ambiguous or open-ended answers. Finally, the alternative-representation experiment is a matched test of auxiliary representations inside one pipeline, isolating representation choice rather than comparing complete end-to-end implementations.

\paragraph{Static and online memory.}
The main experiments construct anchors from completed histories because both benchmarks are evaluated offline, while the update rules themselves are incremental. In a diagnostic on one 19-session LoCoMo conversation, the host and anchors are updated only with sessions available at cutoffs 4, 8, 11, 15, and 19. Anchoring matches the host at the earliest cutoff and improves it by 2.9--6.0\% thereafter, demonstrating causal prefix-only construction (Appendix~\ref{app:streaming}). A full online evaluation would additionally model drift, correction, deletion, and later evidence that revises earlier consolidated records, together with explicit retention policies and provenance-preserving edits.

%% file: sections/conclusion.tex
\section{Conclusion}
\label{sec:conclusion}

Long-horizon memory requires relevant fragments to be retrieved and their relationships to be organized for generation. \gravity{} implements this view as a host-independent entity--event--topic store that consolidates raw dialogue and attaches through query and prompt interfaces. Its gains transfer across five hosts, two benchmarks, two model configurations, and three evaluators, while controlled diagnostics separate organization from full-history consolidation and show an advantage over six matched auxiliary representations. The results support structured anchoring as a portable complement to existing memory systems, with host interaction determined by the evidence each source contributes.

%% file: sections/appendix.tex
\section{Technical appendices and supplementary material}
\label{app}

\subsection{Detailed Introduction of Datasets, Metrics, and Baselines}
\label{app:datasets}

\paragraph{LoCoMo.}
LoCoMo~\citep{maharana2024evaluating} is a benchmark for evaluating very long-term conversational memory of LLM agents.
It contains 10 machine-generated, human-verified multi-session conversations between pairs of speakers, with about 600 turns and 16K tokens per conversation on average and up to 32 sessions.
The conversations cover everyday social topics such as hobbies, work, family, and travel.
The full QA set contains 1{,}986 examples. Following our evaluation protocol, we report accuracy on its 1{,}540 non-adversarial questions: 282 multi-hop questions (Cat\,1, 18.3\%), 841 single-hop questions (Cat\,2, 54.6\%), 321 temporal questions (Cat\,3, 20.8\%), and 96 open-domain questions (Cat\,4, 6.2\%). The remaining 446 adversarial questions (Cat\,5) are excluded.

\paragraph{LongMemEval.}
LongMemEval~\citep{wu2025longmemeval} is a benchmark designed to evaluate five core long-term memory abilities of chat assistants: information extraction, multi-session reasoning, temporal reasoning, knowledge update, and abstention.
It contains 500 carefully designed questions across six base task types, with 30 questions converted into false-premise abstention cases:
\begin{itemize}[leftmargin=1.2em, itemsep=1pt]
    \item \textbf{Single-Session-User (SSU)}: factual recall from user utterances within a single session.
    \item \textbf{Single-Session-Assistant (SSA)}: recall of assistant-generated content.
    \item \textbf{Single-Session-Preference (SSP)}: identifying user preferences expressed in a single session.
    \item \textbf{Multi-Session (MS)}: reasoning across information scattered over multiple sessions.
    \item \textbf{Temporal Reasoning (TR)}: answering questions that require understanding temporal order or duration.
    \item \textbf{Knowledge Update (KU)}: tracking how facts change over time (e.g., the user changed jobs).
    \item \textbf{Abstention (AB)}: correctly declining to answer on the 30 false-premise cases derived from the six base types.
\end{itemize}
Each question is embedded in a scalable multi-session history; we use the benchmark-provided histories for all 500 questions.
We report \emph{overall accuracy} across all 500 questions and separately report accuracy on the 30 abstention cases. In per-task tables, the six base-type columns exclude these 30 cases so that abstention behavior remains visible rather than being folded into the originating task types.

\paragraph{Evaluation metric.}
The primary evaluation uses GPT-4o-mini as the LLM judge, following the original benchmark protocol.
The judge prompt contains the original question, the ground-truth reference answer, and the model's prediction.
The judge outputs a binary label (correct or incorrect) along with a brief justification, from which we compute accuracy.
The cross-model evaluation uses GPT-5.6 sol with the same prompt template, providing an independent-family assessment of the full host--benchmark matrix.

\paragraph{Baseline systems.}
We evaluate five state-of-the-art memory systems that represent distinct architectural paradigms:

\begin{itemize}[leftmargin=1.2em, itemsep=2pt]
    \item \textbf{LightMem}~\citep{fang2026lightmem}: employs a three-stage pipeline of lightweight sensory compression, topic-aware short-term memory construction, and sleep-time long-term memory consolidation. Its long-term memory is retrieved by embedding similarity.
    
    \item \textbf{A-Mem}~\citep{xu2025amem}: follows the Zettelkasten principle to generate atomic notes with contextual descriptions, keywords, tags, and links to related memories. New notes can also update the contextual attributes of existing notes; retrieval is based on embedding similarity.
    
    \item \textbf{Mem0}~\citep{chhikara2025mem0}: extracts and consolidates salient conversational memories in a vector-indexed store. Its graph-enhanced variant additionally represents entities and relations for structured retrieval.
    
    \item \textbf{LiCoMemory}~\citep{huang2025licomemory}: constructs CogniGraph, a lightweight hierarchy whose entities and relations provide semantic indexing layers. Retrieval combines temporal and hierarchy-aware search with reranking.
    
    \item \textbf{ZEP (Graphiti)}~\citep{rasmussen2025zep}: builds a temporal knowledge graph consisting of episodic, semantic, and community subgraphs. Graphiti extracts entities and relationships with temporal metadata, supporting time-aware graph queries alongside vector retrieval.
\end{itemize}

For all systems, we use their default configurations as reported in the original papers or official repositories.
Retrieval limits and fairness controls are detailed in the next section. The code release is available at \url{https://anonymous.4open.science/r/GRAVITY-B217/}.

\subsection{Detailed Experimental Settings}
\label{app:settings}

\paragraph{LLM backbone.}
The primary experiments use GPT-4o-mini for host memory construction, answer generation, anchor extraction, and LLM-judge evaluation, following the original controlled protocol.
The cross-model evaluation changes both sides of this setup: Claude Sonnet 4.6 performs answer generation and anchor extraction, while GPT-5.6 sol serves as an independent cross-family judge. The same generator--judge pairing and prompt templates are used for every host and ablation configuration in this evaluation.
For the Qwen-3-8B anchor variant, Qwen-3-8B is served locally via vLLM on a single NVIDIA H20 GPU; the host systems, answer generator, and judge remain GPT-4o-mini.

\paragraph{Retrieval and fairness.}
The retrieval limit is fixed at 60 and 20 memory entries per query on LoCoMo and LongMemEval respectively, consistent with the default setting in LightMem.
When \gravity{} is attached, the host system still retrieves the same number of entries using its original query.
Expanded queries generated by anchor modules (up to 9, assembled via round-robin interleaving across Topic, Entity, and Event modules) are then submitted to the host's vector search.
The newly retrieved entries replace the 9 lowest-similarity entries from the original retrieval set, keeping the total unchanged.
This design ensures a fair comparison: the host always sees the same number of memory entries, but with potentially broader coverage.
Anchor context (structured blocks for entities, events, and topics) is injected as an \emph{additional} section in the generation prompt, separate from the retrieved memories.

\paragraph{Anchor construction.}
\gravity{} anchors are built offline from raw conversation utterances in fixed-size batches.
The default batch size is $B{=}60$ for Entity and Event modules, and $B{=}150$ for the Topic module.
Each module runs independently via separate LLM calls in the ``default'' configuration; alternatively, the \emph{triple extraction} variant produces Entity, Event, and Topic outputs in a single combined LLM call per batch, reducing build prompt tokens by roughly 75\% at a small accuracy cost (Section~\ref{sec:efficiency}).
The resulting knowledge bases are persisted as portable JSON files and reused across all host systems without modification.

\paragraph{Anchor retrieval.}
At inference time, each anchor module first retrieves candidate entries via native matching: text matching for entities, participant/keyword matching for events, and keyword/label matching for topics.
All candidates are then reranked by cosine similarity between the query embedding and the embedding of each entry's compact text representation.
The top-$K$ entries per module (default $K{=}5$ for each of Entity, Event, and Topic) are retained, subject to a minimum similarity threshold $\sigma$.

\paragraph{Anchor injection format.}
Selected anchors are injected as three formatted context blocks (Topic Summaries, Entity Profiles, Event Records) into the generation prompt.
Each anchor entry is injected in its full structured form:
\begin{itemize}[leftmargin=1.2em, itemsep=1pt]
    \item \textbf{Entity}: canonical name, entity type, all attributes (sorted by confidence), up to 5 relations, up to 5 recent timeline events, and a natural-language summary.
    \item \textbf{Event}: event type label, free-text description, complete 4W1O tuple (Who, What, When, Where, Outcome), recording timestamp, and trace identifier.
    \item \textbf{Topic}: topic label, participants, temporal span, narrative summary, up to 5 key facts, and keywords.
\end{itemize}
The generation prompt instructs the LLM to treat retrieved host memories as the primary source of truth, using anchor context as supplementary structured knowledge for disambiguation and gap-filling.

\subsection{Detailed Experimental Results}

\subsubsection{Main Results}
\label{app:main_results}

Tables~\ref{tab:main_locomo_detail} and~\ref{tab:main_longmemeval_detail} present the full per-category and per-task breakdowns of the primary GPT-4o-mini results reported in Table~\ref{tab:main}.

\begin{table*}[h]
\centering
\caption{Primary per-category results on LoCoMo under the GPT-4o-mini configuration (LLM-judge accuracy, \%). Cat\,1: multi-hop, Cat\,2: single-hop, Cat\,3: temporal, Cat\,4: open-domain.}
\label{tab:main_locomo_detail}
\small
\setlength{\tabcolsep}{6pt}
\renewcommand{\arraystretch}{1.12}
\begin{tabular}{@{}l ccccc ccccc c@{}}
\toprule
& \multicolumn{5}{c}{\textbf{w/o Anchor}} & \multicolumn{5}{c}{\textbf{w/ Anchor}} & \\
\cmidrule(lr){2-6} \cmidrule(lr){7-11}
\textbf{Host System}
  & \textbf{Cat\,1} & \textbf{Cat\,2} & \textbf{Cat\,3} & \textbf{Cat\,4} & \textbf{All}
  & \textbf{Cat\,1} & \textbf{Cat\,2} & \textbf{Cat\,3} & \textbf{Cat\,4} & \textbf{All}
  & \textbf{$\Delta$} \\
\midrule
LightMem     & 60.6 & 75.9 & 70.7 & 45.8 & 70.1 & 65.3 & 82.1 & 76.9 & 47.9 & 75.8 & {+5.7} \\
A-Mem        & 55.0 & 73.7 & 60.4 & 38.5 & 65.3 & 63.5 & 79.2 & 63.6 & 44.8 & 70.9 & {+5.6} \\
Mem0         & 51.8 & 56.2 & 40.2 & 38.5 & 51.0 & 55.7 & 65.2 & 52.0 & 39.6 & 59.1 & {+8.1} \\
LiCoMemory   & 46.5 & 61.1 & 53.6 & 41.7 & 55.7 & 58.2 & 73.7 & 59.8 & 45.8 & 66.2 & {+10.5} \\
ZEP          & 51.4 & 57.9 & 52.7 & 42.7 & 54.7 & 58.2 & 66.9 & 57.0 & 43.8 & 61.8 & {+7.1} \\
\bottomrule
\end{tabular}
\end{table*}

\begin{table*}[h]
\centering
\caption{Primary per-task results on LongMemEval under the GPT-4o-mini configuration (accuracy, \%). The six base-type columns exclude the 30 false-premise cases; AB reports those cases separately. Overall is accuracy across all 500 questions.}
\label{tab:main_longmemeval_detail}
\small
\setlength{\tabcolsep}{8pt}
\renewcommand{\arraystretch}{1.12}
\begin{tabular}{@{}l ccccccc c c@{}}
\toprule
& \multicolumn{7}{c}{\textbf{Per-Task Accuracy}} & \textbf{Aggregate} & \\
\cmidrule(lr){2-8} \cmidrule(lr){9-9}
\textbf{Host System}
  & \textbf{SSU} & \textbf{SSA} & \textbf{SSP} & \textbf{MS} & \textbf{TR} & \textbf{KU} & \textbf{AB}
  & \textbf{Overall} & \textbf{$\Delta$} \\
\midrule
\multicolumn{10}{@{}l}{\textit{w/o Anchor}} \\
LightMem     & 87.1 & 32.1 & 68.2 & 71.7 & 67.2 & 83.1 & 60.0 & 68.8 & -- \\
A-Mem        & 96.9 & 92.9 & 13.3 & 39.7 & 42.5 & 61.1 & 16.7 & 53.8 & -- \\
Mem0         & 81.3 & 33.9 & 10.0 & 38.0 & 58.3 & 63.9 & 76.7 & 52.6 & -- \\
LiCoMemory   & 92.2 & 87.5 & 43.3 & 35.5 & 41.7 & 76.4 & 50.0 & 57.4 & -- \\
ZEP          & 37.5 & 51.8 & 26.7 & 38.8 & 57.5 & 55.6 & 66.7 & 48.2 & -- \\
\midrule
\multicolumn{10}{@{}l}{\textit{w/ Anchor}} \\
LightMem     & 100.0 & 33.9 & 76.7 & 73.6 & 68.5 & 90.3 & 53.3 & 72.6 & {+3.8} \\
A-Mem        & 95.3 & 98.2 & 23.3 & 53.7 & 52.8 & 61.1 & 56.7 & 63.2 & {+9.4} \\
Mem0         & 93.8 & 28.6 & 70.0 & 54.6 & 70.9 & 66.7 & 63.3 & 64.0 & {+11.4} \\
LiCoMemory   & 95.3 & 89.3 & 76.7 & 50.4 & 53.5 & 72.2 & 60.0 & 66.6 & {+9.2} \\
ZEP          & 64.1 & 60.7 & 60.0 & 56.2 & 59.8 & 65.3 & 60.0 & 60.4 & {+12.2} \\
\bottomrule
\end{tabular}
\end{table*}

\subsubsection{Cross-Model Evaluation}
\label{app:cross_model}

We evaluate the complete five-host, two-benchmark matrix under a second model configuration, with Claude Sonnet 4.6 for answer generation and anchor extraction and GPT-5.6 sol as the judge. This changes both the base model and evaluation model relative to the primary GPT-4o-mini configuration. Table~\ref{tab:cross_model_main} in the main paper reports the overall results; Tables~\ref{tab:cross_model_locomo_detail} and~\ref{tab:cross_model_lme_detail} provide the category-level breakdowns.

\begin{table*}[h]
\centering
\caption{Cross-model evaluation on LoCoMo (LLM-judge accuracy, \%).}
\label{tab:cross_model_locomo_detail}
\small
\setlength{\tabcolsep}{6pt}
\begin{tabular}{@{}l ccccc ccccc c@{}}
\toprule
& \multicolumn{5}{c}{\textbf{w/o Anchor}} & \multicolumn{5}{c}{\textbf{w/ Anchor}} & \\
\cmidrule(lr){2-6} \cmidrule(lr){7-11}
\textbf{Host} & \textbf{Cat\,1} & \textbf{Cat\,2} & \textbf{Cat\,3} & \textbf{Cat\,4} & \textbf{All} & \textbf{Cat\,1} & \textbf{Cat\,2} & \textbf{Cat\,3} & \textbf{Cat\,4} & \textbf{All} & \textbf{$\Delta$} \\
\midrule
LightMem   & 69.5 & 82.0 & 79.8 & 58.3 & 77.8 & 71.3 & 90.6 & 84.7 & 59.4 & 83.9 & {+6.1} \\
A-Mem      & 64.5 & 77.1 & 66.4 & 55.2 & 71.2 & 69.9 & 90.1 & 87.2 & 58.3 & 83.8 & {+12.6} \\
Mem0       & 45.4 & 59.2 & 48.3 & 42.7 & 53.4 & 70.5 & 85.0 & 78.2 & 59.4 & 79.3 & {+25.9} \\
LiCoMemory & 51.8 & 63.0 & 53.9 & 47.9 & 58.1 & 68.4 & 86.7 & 82.2 & 58.3 & 80.7 & {+22.6} \\
ZEP        & 51.4 & 60.0 & 53.6 & 51.0 & 56.6 & 70.9 & 85.1 & 79.8 & 59.4 & 79.8 & {+23.2} \\
\bottomrule
\end{tabular}
\end{table*}

\begin{table*}[h]
\centering
\caption{Cross-model evaluation on LongMemEval (accuracy, \%). The six base-type columns exclude AB; Overall covers all 500 questions.}
\label{tab:cross_model_lme_detail}
\small
\setlength{\tabcolsep}{8pt}
\begin{tabular}{@{}l ccccccc c c@{}}
\toprule
\textbf{Host} & \textbf{SSU} & \textbf{SSA} & \textbf{SSP} & \textbf{MS} & \textbf{TR} & \textbf{KU} & \textbf{AB} & \textbf{Overall} & \textbf{$\Delta$} \\
\midrule
\multicolumn{10}{@{}l}{\textit{w/o Anchor}} \\
LightMem   & 92.2 & 23.2 & 70.0 & 68.6 & 74.0 & 79.2 & 83.3 & 70.4 & -- \\
A-Mem      & 96.9 & 100.0 & 46.7 & 68.6 & 85.8 & 84.7 & 90.0 & 82.4 & -- \\
Mem0       & 84.4 & 10.7 & 60.0 & 66.1 & 73.2 & 79.2 & 86.7 & 66.8 & -- \\
LiCoMemory & 90.6 & 94.6 & 80.0 & 53.7 & 63.0 & 84.7 & 70.0 & 72.4 & -- \\
ZEP        & 40.6 & 51.8 & 43.3 & 50.4 & 74.0 & 58.3 & 76.7 & 57.6 & -- \\
\midrule
\multicolumn{10}{@{}l}{\textit{w/ Anchor}} \\
LightMem   & 98.4 & 42.9 & 70.0 & 83.5 & 88.2 & 93.1 & 96.7 & 83.4 & {+13.0} \\
A-Mem      & 100.0 & 98.2 & 86.7 & 79.3 & 81.9 & 87.5 & 90.0 & 87.0 & {+4.6} \\
Mem0       & 95.3 & 30.4 & 86.7 & 81.8 & 89.0 & 84.7 & 93.3 & 81.0 & {+14.2} \\
LiCoMemory & 98.4 & 89.3 & 96.7 & 79.3 & 88.2 & 91.7 & 86.7 & 88.4 & {+16.0} \\
ZEP        & 93.8 & 67.9 & 86.7 & 77.7 & 92.9 & 90.3 & 93.3 & 85.8 & {+28.2} \\
\bottomrule
\end{tabular}
\end{table*}

\begin{figure}[h]
\centering
\includegraphics[width=\linewidth]{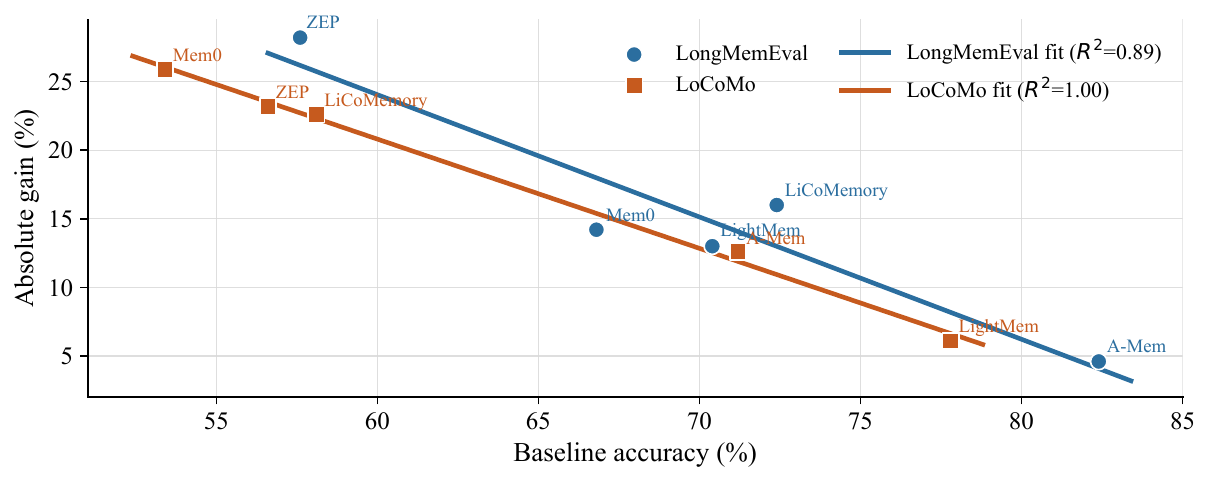}
\caption{Gain versus baseline strength in the cross-model evaluation. Each point is one host--benchmark pair; solid lines show per-benchmark regressions.}
\label{fig:gain-baseline}
\end{figure}

\paragraph{Gain--Baseline Relationship.}
\label{sec:gain-baseline}
The cross-model evaluation also shows that weaker hosts receive larger gains. Across its 10 host--benchmark points, the pooled regression yields a slope of $-0.79$, $R^2 = 0.92$, and $p < 0.0001$ (Figure~\ref{fig:gain-baseline}). The trend holds separately on LongMemEval (slope~$= -0.89$, $R^2 = 0.89$, $p = 0.015$) and LoCoMo (slope~$= -0.80$, $R^2 = 1.00$, $p < 0.001$).

This relationship is consistent with larger benefits when the host baseline leaves more headroom. The anchor-only and oracle diagnostics below show that the gain combines two effects: independent access to structured information from the full history and improved organization of evidence already available to the generator. The stronger average gains in this configuration also show that effect size depends on the generator and judge pairing, while the direction remains stable across both evaluated configurations.

\paragraph{Evaluator sensitivity on fixed outputs.}
To isolate evaluator choice, we keep the answers from the primary GPT-4o-mini runs unchanged and score them with GPT-4o-mini, Claude Sonnet 4.6, and GPT-5.6 sol. Table~\ref{tab:judge_robustness} therefore compares the three evaluators on identical candidate answers rather than new generations. All 30 host--benchmark--evaluator deltas remain positive. The average gains are 9.2\%/9.5\%/8.3\% on LongMemEval and 7.5\%/6.4\%/6.2\% on LoCoMo, respectively.

\begin{table*}[h]
\centering
\caption{Evaluator sensitivity using identical outputs from the primary GPT-4o-mini runs. Each cell reports Base$\to$+\gravity{} ($\Delta$), in \%.}
\label{tab:judge_robustness}
\scriptsize
\setlength{\tabcolsep}{2.7pt}
\renewcommand{\arraystretch}{1.08}
\begin{tabular}{@{}l ccc ccc@{}}
\toprule
& \multicolumn{3}{c}{\textbf{LongMemEval}} & \multicolumn{3}{c}{\textbf{LoCoMo}} \\
\cmidrule(lr){2-4} \cmidrule(lr){5-7}
\textbf{Host} & \textbf{GPT-4o-mini} & \textbf{Sonnet 4.6} & \textbf{GPT-5.6 sol} & \textbf{GPT-4o-mini} & \textbf{Sonnet 4.6} & \textbf{GPT-5.6 sol} \\
\midrule
ZEP        & 48.2$\to$60.4 (+12.2) & 48.6$\to$61.0 (+12.4) & 52.4$\to$61.2 (+8.8) & 54.7$\to$61.8 (+7.1) & 57.1$\to$63.3 (+6.2) & 55.6$\to$62.1 (+6.6) \\
Mem0       & 52.6$\to$64.0 (+11.4) & 53.2$\to$64.6 (+11.4) & 54.2$\to$65.2 (+11.0) & 51.0$\to$59.1 (+8.1) & 51.9$\to$57.9 (+6.0) & 54.2$\to$61.3 (+7.1) \\
LiCoMemory & 57.4$\to$66.6 (+9.2)  & 56.6$\to$67.4 (+10.8) & 59.8$\to$68.4 (+8.6) & 55.7$\to$66.2 (+10.5) & 57.6$\to$67.9 (+10.3) & 57.9$\to$67.7 (+9.7) \\
A-Mem      & 53.8$\to$63.2 (+9.4)  & 54.0$\to$62.0 (+8.0)  & 55.8$\to$63.8 (+8.0) & 65.3$\to$70.9 (+5.6) & 64.4$\to$69.1 (+4.7) & 66.6$\to$71.9 (+5.3) \\
LightMem   & 68.8$\to$72.6 (+3.8)  & 67.2$\to$72.0 (+4.8)  & 67.0$\to$72.2 (+5.2) & 70.1$\to$75.8 (+5.7) & 69.3$\to$74.1 (+4.7) & 73.8$\to$76.2 (+2.5) \\
\midrule
\textit{Average $\Delta$} & +9.2 & +9.5 & +8.3 & +7.5 & +6.4 & +6.2 \\
\bottomrule
\end{tabular}
\end{table*}

\paragraph{Mechanism decomposition and evidence coverage.}
\label{app:mechanism}
The controlled decomposition in Table~\ref{tab:controlled}(b) uses LightMem on full LoCoMo with Claude Sonnet 4.6 generation and anchor extraction and GPT-5.6 sol evaluation. The same-evidence condition first runs the host retrieval, then builds query-specific entity, event, and topic records only from the sessions represented in that retrieved context. Query expansion is disabled, so the generator receives no source session outside the host-only prompt. The full-history condition changes only the construction scope to the complete conversation.

Two additional replacement controls clarify why the two memory sources are not interchangeable. Replacing the host entries with records built from its retrieved sessions yields 77.1\%; replacing them with records built from the full history yields 80.3\%. In both cases, removing the raw host entries reduces accuracy by about 5.2\% relative to keeping them, consistent with the observed losses from over-summarization and exact-detail deletion.

We also measure retrieval-side coverage without generation or an LLM judge. LoCoMo provides gold evidence turns for each question. Mapping host entries and anchor provenance to source sessions, host retrieval attains 0.945 mean gold-session recall and fully covers the gold sessions for 0.900 of questions. Adding anchors increases these values to 0.977 and 0.960. This shows that the external bank reaches labeled evidence that the host did not retrieve, while the same-evidence condition separately measures the organization benefit.

\paragraph{Alternative structured representations.}
Table~\ref{tab:controlled}(a) in the main paper compares six auxiliary representations under a fixed LightMem pipeline. The prior-system rows adapt their representation and aligned retrieval procedure; they are not end-to-end reproductions of the original systems. The entity--event--topic representation reaches 83.9\%, 3.6\% above the strongest alternative, while a free-form summary improves the base by only 0.2\%.

\subsubsection{Ablation Results}
\label{app:ablation}

\begin{table}[h]
\centering
\caption{Primary per-category ablation on LoCoMo under the GPT-4o-mini configuration (LLM-judge accuracy, \%). Cat\,1: multi-hop, Cat\,2: single-hop, Cat\,3: temporal, Cat\,4: open-domain. All variants use LightMem as the host system.}
\label{tab:ablation_locomo_detail}
\small
\setlength{\tabcolsep}{18pt}
\renewcommand{\arraystretch}{1.12}
\begin{tabular}{@{}l ccccc@{}}
\toprule
\textbf{Configuration} & \textbf{Cat\,1} & \textbf{Cat\,2} & \textbf{Cat\,3} & \textbf{Cat\,4} & \textbf{Overall} \\
\midrule
LightMem (baseline) & 60.6 & 75.9 & 70.7 & 45.8 & 70.1 \\
\quad + Summary (unstructured) & 61.3 & 77.9 & 70.7 & 46.9 & 71.4 \\
\midrule
\multicolumn{6}{@{}l}{\textit{Single module}} \\
\quad + Entity only     & 68.8 & 78.8 & 65.7 & 47.9 & 72.3 \\
\quad + Event only      & 63.5 & 76.0 & 69.8 & 50.0 & 70.8 \\
\quad + Topic only      & 63.1 & 79.7 & 70.4 & 49.0 & 72.8 \\
\midrule
\multicolumn{6}{@{}l}{\textit{Pairwise}} \\
\quad + Entity + Event  & 63.8 & 78.0 & 69.8 & 52.1 & 72.1 \\
\quad + Entity + Topic  & 67.0 & 80.3 & 71.7 & 46.9 & 74.0 \\
\quad + Event + Topic   & 65.6 & 80.7 & 71.7 & 46.9 & 74.0 \\
\midrule
\rowcolor{gray!8}
+ All three (full \gravity{}) & 65.3 & \textbf{82.1} & \textbf{76.9} & 47.9 & \textbf{75.8} \\
\midrule
\multicolumn{6}{@{}l}{\textit{Design choices}} \\
\quad $-$ query expansion  & 65.3 & \textbf{82.2} & 74.1 & 49.0 & 75.3 \\
\quad $-$ reranking        & 64.2 & 78.6 & 70.4 & 49.0 & 72.4 \\
\bottomrule
\end{tabular}
\end{table}

Tables~\ref{tab:ablation_locomo_detail} and~\ref{tab:ablation_longmemeval_detail} present the full per-category and per-task ablation breakdowns, complementing the compact summary in Table~\ref{tab:ablation}.

\paragraph{LoCoMo per-category analysis (Table~\ref{tab:ablation_locomo_detail}).}
Entity gives the strongest single-module gain on multi-hop questions (60.6$\to$68.8\%), Event gives the largest single-module improvement on open-domain questions (45.8$\to$50.0\%), and Topic gives the strongest single-module overall result (72.8\%).
The full combination reaches the best overall accuracy (75.8\%) and the best single-hop and temporal accuracy, supporting complementarity among the three views.
Removing reranking lowers overall accuracy to 72.4\%, whereas removing query expansion retains 75.3\%, indicating that anchor selection is more important than expanded retrieval in this configuration.

\paragraph{LongMemEval per-task analysis (Table~\ref{tab:ablation_longmemeval_detail}).}

\begin{table*}[h]
\centering
\caption{Primary per-task ablation on LongMemEval under the GPT-4o-mini configuration (accuracy, \%). The six base-type columns contain 470 non-abstention questions (64/56/30/121/127/72); AB contains the 30 false-premise cases. Overall is accuracy across all 500 questions. All variants use LightMem as host.}
\label{tab:ablation_longmemeval_detail}
\small
\setlength{\tabcolsep}{8.5pt}
\renewcommand{\arraystretch}{1.12}
\begin{tabular}{@{}l ccccccc c@{}}
\toprule
& \multicolumn{7}{c}{\textbf{Per-Task Accuracy}} & \textbf{Aggregate} \\
\cmidrule(lr){2-8} \cmidrule(lr){9-9}
\textbf{Configuration}
  & \textbf{SSU} & \textbf{SSA} & \textbf{SSP} & \textbf{MS} & \textbf{TR} & \textbf{KU} & \textbf{AB}
  & \textbf{Overall} \\
\midrule
LightMem (baseline)
  & 87.1 & 32.1 & 68.2 & 71.7 & 67.2 & 83.1 & 60.0 & 68.8 \\
\quad + Summary (unstructured)
  & 90.6 & 30.4 & 73.3 & 71.7 & 67.8 & 87.5 & 53.3 & 69.0 \\
\midrule
\multicolumn{9}{@{}l}{\textit{Single module}} \\
\quad + Entity only
  & 92.2 & 25.0 & 70.0 & 76.9 & 70.1 & 90.3 & 56.7 & 71.6 \\
\quad + Event only
  & 93.8 & 28.6 & 63.3 & 76.0 & 68.5 & 86.1 & 56.7 & 70.6 \\
\quad + Topic only
  & 96.9 & 32.1 & 73.3 & 70.3 & 67.7 & 87.5 & 46.7 & 70.0 \\
\midrule
\multicolumn{9}{@{}l}{\textit{Pairwise}} \\
\quad + Entity + Event
  & 93.8 & 26.8 & 66.7 & 74.4 & 66.9 & 88.9 & 53.3 & 70.0 \\
\quad + Entity + Topic
  & 95.3 & 32.1 & \textbf{80.0} & 74.4 & 65.4 & 87.5 & 53.3 & 71.0 \\
\quad + Event + Topic
  & 93.8 & 32.1 & 76.7 & 71.1 & 64.6 & 84.7 & 50.0 & 69.0 \\
\midrule
\rowcolor{gray!8}
+ All three (full \gravity{})
  & \textbf{100.0} & 33.9 & 76.7 & 73.6 & 68.5 & 90.3 & 53.3 & 72.6 \\
\midrule
\multicolumn{9}{@{}l}{\textit{Design choices}} \\
\quad $-$ query expansion
  & 95.3 & 33.9 & \textbf{80.0} & \textbf{75.2} & \textbf{70.9} & \textbf{91.7} & 46.7 & \textbf{73.0} \\
\quad $-$ reranking
  & 95.3 & \textbf{35.7} & 56.7 & 68.6 & 65.4 & 87.5 & \textbf{56.7} & 68.8 \\
\bottomrule
\end{tabular}
\end{table*}

The full \gravity{} configuration (+EVT) raises overall accuracy from 68.8\% to 72.6\%. It gives the strongest SSU result (100.0\%) and improves SSP, MS, TR, and KU, while AB decreases from 60.0\% to 53.3\%.
No single module dominates across tasks: Entity is strongest on MS and KU, Event improves MS, and Topic is strongest on SSP among the single-module variants.
Removing query expansion yields the highest overall accuracy (73.0\%), whereas removing reranking returns overall accuracy to the 68.8\% baseline. This pattern shows that structured anchor injection supplies the main benefit and reranking is required to control anchor noise.

\subsubsection{Efficiency and Build Variants}
\label{app:efficiency}

\begin{table}[h]
\centering
\caption{Build-phase cost per conversation under the primary GPT-4o-mini configuration. Tokens are reported in thousands; time is end-to-end wall-clock seconds.}
\label{tab:efficiency}
\footnotesize
\setlength{\tabcolsep}{4pt}
\begin{tabular}{@{}l rrr r rrr r@{}}
\toprule
& \multicolumn{4}{c}{\textbf{LoCoMo}} & \multicolumn{4}{c}{\textbf{LongMemEval}} \\
\cmidrule(lr){2-5} \cmidrule(lr){6-9}
\textbf{System} & \textbf{In} & \textbf{Out} & \textbf{Total} & \textbf{Time} & \textbf{In} & \textbf{Out} & \textbf{Total} & \textbf{Time} \\
\midrule
LightMem & 91.9 & 22.2 & 114.1 & 848.5 & 98.3 & 13.4 & 111.7 & 496.0 \\
A-Mem & 912.6 & 236.8 & 1149.4 & 6060.7 & 1372.2 & 233.6 & 1605.8 & 5132.0 \\
Mem0 & 1483.4 & 210.0 & 1693.4 & 4432.9 & 984.3 & 168.3 & 1152.6 & 4248.5 \\
LiCoMemory & 302.9 & 48.2 & 351.1 & 901.7 & 460.4 & 129.1 & 590.5 & 1803.3 \\
ZEP & 733.0 & 13.7 & 746.6 & 780.9 & 1247.9 & 111.0 & 1358.9 & 1275.9 \\
\midrule
\rowcolor{gray!8}\gravity{} anchors & 152.4 & 40.2 & 192.6 & 556.8 & 90.6 & 27.2 & 117.8 & 360.6 \\
\bottomrule
\end{tabular}
\end{table}

\begin{table}[h]
\centering
\caption{Anchor build variants on LoCoMo under the primary configuration. Parallel uses concurrent module calls; Triple fuses the three extractors; Qwen replaces GPT-4o-mini only for anchor construction.}
\label{tab:build-variants}
\footnotesize
\setlength{\tabcolsep}{3.5pt}
\begin{tabular}{@{}l ccccc@{}}
\toprule
\textbf{Host} & \textbf{No Anchor} & \textbf{Default} & \textbf{Parallel} & \textbf{Triple} & \textbf{Qwen} \\
\midrule
LightMem & 70.1 & \textbf{75.8} & \textbf{75.8} & 74.2 & 72.7 \\
Mem0 & 51.0 & \textbf{59.1} & \textbf{59.1} & 55.0 & 57.3 \\
A-Mem & 65.3 & \textbf{70.9} & \textbf{70.9} & 68.7 & 69.5 \\
LiCoMemory & 55.7 & \textbf{66.2} & \textbf{66.2} & 64.4 & \textbf{66.2} \\
ZEP & 54.7 & 61.8 & 61.8 & 58.0 & \textbf{62.5} \\
\midrule
\textit{Build Tokens (K)} & -- & 193 & 193 & \textbf{48} & 286 \\
\textit{Build Time (s)} & -- & 557 & \textbf{280} & 453 & 667 \\
\bottomrule
\end{tabular}
\end{table}

The default anchor build is shared by all hosts. Parallel execution changes latency but not predictions. Triple extraction reduces token use by 75\% and lowers average accuracy by 1.6\%; the open-weight Qwen-3-8B variant removes API cost while remaining competitive on several hosts.

\subsubsection{Parameter Sensitivity}
\label{app:param}

Under the primary GPT-4o-mini configuration, we examine the sensitivity of \gravity{} to its key hyperparameters on a representative LoCoMo conversation (conv-42, 199 questions, LightMem host). Parameters fall into two groups: \emph{build-phase} parameters that control extraction granularity and cost, and \emph{inference-phase} parameters that control how much anchor content reaches the generator.

\begin{figure}[h]
\centering
\includegraphics[width=\linewidth]{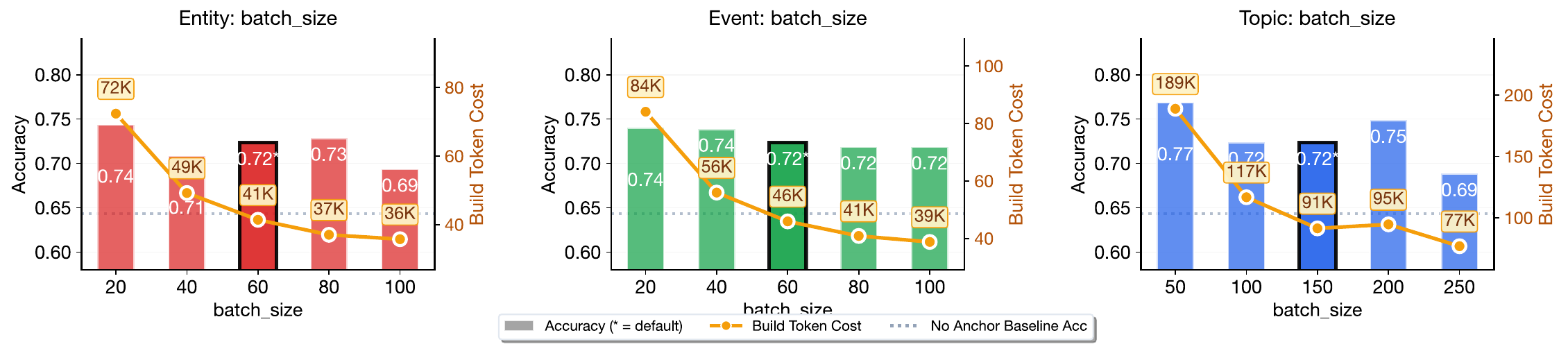}
\caption{Build-phase parameter sensitivity: batch size vs.\ accuracy (bars, left axis) and build token cost (line, right axis) for Entity, Event, and Topic modules. Dashed line: no-anchor baseline. Default values marked with black border.}
\label{fig:param}
\end{figure}

\paragraph{Build-phase: batch size (Figure~\ref{fig:param}).}
Batch size $B$ governs how many utterances are processed per LLM call during anchor extraction.
Across a 5$\times$ range ($B{=}20$ to $100$ for Entity/Event; $B{=}50$ to $250$ for Topic), all configurations outperform the no-anchor baseline, with accuracy varying by less than 6\%.
Smaller batches yield finer-grained extraction at higher token cost: the smallest batch ($B{=}20$) achieves the highest Entity accuracy (74.4\%) but costs 72K tokens, nearly double the default.
The default settings ($B{=}60$ for Entity/Event, $B{=}150$ for Topic) sit at the knee of the cost--performance curve, achieving within 1--2\% of peak accuracy at roughly half the token cost.

\emph{Takeaway}: anchor quality is robust to batch size; the default balances accuracy and cost without requiring per-dataset tuning.

\begin{figure}[h]
\centering
\includegraphics[width=0.8\linewidth]{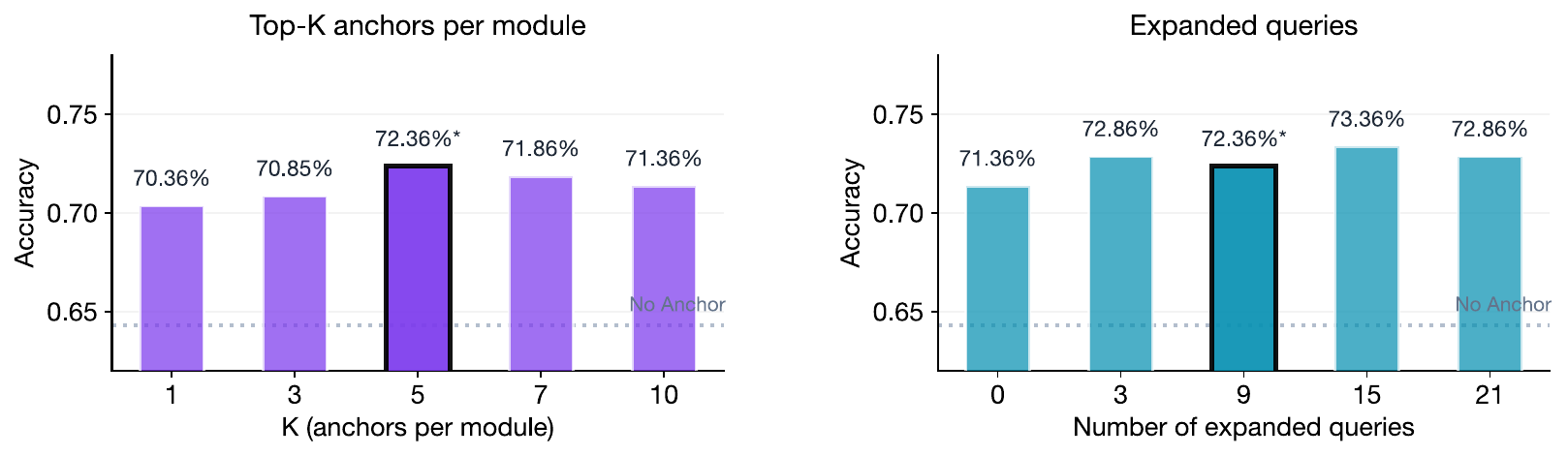}
\caption{Inference-phase parameter sensitivity. \textbf{Left}: top-$K$ anchors per module. \textbf{Right}: number of expanded queries. Dashed line: no-anchor baseline. Default values marked with *.}
\label{fig:param-inference}
\end{figure}

\paragraph{Inference-phase: top-$K$ and query expansion (Figure~\ref{fig:param-inference}).}
Two parameters control how much anchor and expanded retrieval content is injected at inference time: the number of anchors retained per module ($K$), and the number of expanded retrieval queries.

\noindent \emph{Top-K anchors per module.}
Increasing $K$ from 1 to 5 steadily improves accuracy (70.4\%$\to$72.4\%), as the generator gains access to a richer structural context.
Beyond $K{=}5$, accuracy slightly declines (71.9\% at $K{=}7$, 71.4\% at $K{=}10$), indicating that lower-ranked anchors introduce noise that dilutes the signal from the most relevant entries.

\noindent \emph{Number of expanded queries.}
With zero expansion, the system already achieves 71.4\%, confirming that structured context injection alone provides substantial gains.
Adding expanded queries yields a modest further improvement, peaking at 73.4\% with 15 queries.
The gains are relatively flat across 3--21 queries, confirming that query expansion is secondary to structured context injection (\S\ref{sec:ablation}).

\emph{Takeaway}: $K{=}5$ is the sweet spot for anchor density; query expansion provides marginal additional benefit and is insensitive to exact count. Both findings support our default choices without requiring task-specific tuning.

\subsubsection{Inference Latency}
\label{app:latency}

\begin{table}[h]
\centering
\caption{Inference-time latency and token consumption under the primary GPT-4o-mini configuration.
We report the average response time (seconds) and average total tokens per query,
with and without \gravity{} anchors.
$\Delta$ denotes the overhead introduced by the anchor module.}
\label{tab:latency}
\small
\setlength{\tabcolsep}{10.8pt}
\begin{tabular}{ll cc c cc c}
\toprule
& & \multicolumn{3}{c}{\textbf{Avg Time (s) $\downarrow$}} & \multicolumn{3}{c}{\textbf{Avg Tokens $\downarrow$}} \\
\cmidrule(lr){3-5} \cmidrule(lr){6-8}
\textbf{Benchmark} & \textbf{Host System} & w/o & w/ & $\Delta$ & w/o & w/ & $\Delta$ \\
\midrule
\multirow{5}{*}{\textit{LoCoMo}}
 & LightMem    & 1.57 & 2.03 & +0.46 & 2313 & 4211 & +1898 \\
 & A-Mem       & 1.60 & 1.95 & +0.35 & 4691 & 6584 & +1893 \\
 & Mem0        & 1.33 & 1.64 & +0.31 & 3724 & 5881 & +2158 \\
 & LiCoMemory  & 8.27 & 9.19 & +0.92 & 1618 & 3745 & +2127 \\
 & ZEP         & 1.83 & 3.86 & +2.03 & 5474 & 7580 & +2106 \\
\midrule
\multirow{5}{*}{\textit{LongMemEval}}
 & LightMem    & 2.51 & 2.96 & +0.45 &  819 & 1223 &  +404 \\
 & A-Mem       & 2.11 & 2.64 & +0.53 & 5727 & 6128 &  +401 \\
 & Mem0        & 2.80 & 3.45 & +0.65 & 2322 & 2542 &  +220 \\
 & LiCoMemory  & 10.34 & 14.56 & +4.22 & 1592 & 2053 &  +461 \\
 & ZEP         & 2.04 & 4.61 & +2.57 & 6036 & 6508 &  +472 \\
\bottomrule
\end{tabular}
\end{table}

For most hosts, \gravity{} adds less than 1\,s and approximately 2K tokens per query.
The overhead on LongMemEval is smaller in absolute token terms because the retrieval window is shorter (20 vs.\ 60 entries), resulting in a more compact anchor context.
LiCoMemory and ZEP exhibit larger time overheads due to their graph-based retrieval pipelines, which are more sensitive to additional query load from expanded queries.

\subsection{Extended Discussion and Theoretical Proofs}

\subsubsection{Full Results of the Oracle Experiment}
\label{sec:appendix_oracle}

Table~\ref{tab:oracle} provides the complete categorical results for the Oracle experiment discussed in Section~\ref{sec:oracle}. 

\begin{table}[hbtp]
\centering
\caption{Oracle diagnostic under the primary GPT-4o-mini configuration on LoCoMo (A-Mem host, 1{,}540 non-adversarial questions).}

\label{tab:oracle}
\small
\setlength{\tabcolsep}{6pt}
\begin{tabular}{l cc cc cc}
\toprule
& \multicolumn{2}{c}{\textbf{Oracle Only}} & \multicolumn{2}{c}{\textbf{Oracle Front}} & \multicolumn{2}{c}{\textbf{Oracle Random}} \\
\cmidrule(lr){2-3} \cmidrule(lr){4-5} \cmidrule(lr){6-7}
\textbf{Category} & w/o Anchor & w/ Anchor & w/o Anchor & w/ Anchor & w/o Anchor & w/ Anchor \\
\midrule
Multi-hop (C1)   & 79.1 & 78.7 & 74.1 & 77.0 & 70.2 & 72.3 \\
Single-hop (C2)  & 88.3 & 88.8 & 85.1 & 87.4 & 83.8 & 85.9 \\
Temporal (C3)    & 89.1 & 84.4 & 85.0 & 85.0 & 66.4 & 73.8 \\
Open-domain (C4) & 58.3 & 54.2 & 50.0 & 54.2 & 50.0 & 47.9 \\
\midrule
\textbf{Overall}  & \textbf{84.9} & 83.9 & 80.9 & \textbf{82.9} & 75.6 & \textbf{78.5} \\
\bottomrule
\end{tabular}
\end{table}

\subsubsection{A Diminishing-Returns Model}
\label{sec:appendix_proof}

This section gives a simple explanatory model for the diminishing-returns pattern discussed in Section~\ref{sec:oracle}. It characterizes one source of the observed relationship, not a causal proof of the cross-host regression.

Let $P_{success}$ be the probability of a language model answering a long-horizon question correctly, which depends on the density of structured evidence $\rho$ available in the generation context. We model this as an exponential cumulative distribution function:
\begin{equation}
    P_{success} = 1 - e^{-\lambda \rho}
\end{equation}
where $\lambda > 0$ denotes the task-specific decay constant. 

\textbf{Modeling assumption.} The exponential form is a convenient saturating function that keeps $P \in [0,1]$ and captures diminishing returns as additional structured evidence accumulates. It can also be motivated by treating the discovery of a valid reasoning path as a Poisson process in which structural clues contribute independently. We use it as an interpretable approximation rather than asserting that LLM reasoning follows this process exactly.

Let $\rho_{base}$ be the inherent structural density successfully captured and presented by the host system's baseline retrieval. The baseline accuracy is therefore:
\begin{equation}
    P_{base} = 1 - e^{-\lambda \rho_{base}}
\end{equation}

\gravity{} acts as an external plugin that injects additional structured anchoring information, denoted as $\Delta \rho$. The augmented accuracy becomes:
\begin{equation}
    P_{GRAVITY} = 1 - e^{-\lambda (\rho_{base} + \Delta \rho)}
\end{equation}

We derive the absolute accuracy gain $\Delta P$ provided by the structured anchors for a given query as follows:
\begin{align}
    \Delta P &= P_{GRAVITY} - P_{base} \notag \\
             &= e^{-\lambda \rho_{base}} - e^{-\lambda (\rho_{base} + \Delta \rho)} \notag \\
             &= e^{-\lambda \rho_{base}} \left( 1 - e^{-\lambda \Delta \rho} \right)
\end{align}

Notice that $e^{-\lambda \rho_{base}} = 1 - P_{base}$. Substituting this back into the equation, we obtain:
\begin{equation}
    \Delta P = \left( 1 - e^{-\lambda \Delta \rho} \right) \cdot (1 - P_{base})
\end{equation}

For a fixed effective increment $\Delta\rho$, Equation~\ref{eq:final} is linear in baseline headroom:
\begin{equation}
    \Delta P = K(1-P_{base}), \qquad K = 1-e^{-\lambda\Delta\rho}.
    \label{eq:final}
\end{equation}
Thus, holding the usable anchor contribution fixed, a stronger baseline leaves less room for absolute improvement. Across real hosts, however, $K$ need not be constant. Although anchor construction and retrieval are independent of host internals, the generator consumes host evidence and anchors jointly; redundancy, conflict, and ordering can change how much anchor information is effectively usable. Table~\ref{tab:anchor_only} illustrates this interaction: host evidence adds consistently beyond anchors alone on LongMemEval, but its incremental value is mixed on LoCoMo.

Empirically, an unconstrained linear regression over the 10 host--benchmark points yields a slope of $-0.79$, $R^2 = 0.92$, and $p < 0.0001$. This result is consistent with headroom-driven diminishing returns, but it does not establish a host-invariant $K$ or identify a causal mechanism from ten aggregate observations.

\paragraph{Connecting aggregate trends to query-level variation.}
The low pairwise Jaccard similarity ($0.09$--$0.17$) among per-host gain sets shows that different hosts benefit on different questions. This is compatible with the model above when $\Delta\rho$ varies by query and host context. Together with the anchor-only diagnostic, the evidence supports two effects: the independent anchor store can recover and organize information on its own, while joint prompting can either add complementary host evidence or introduce redundancy. Architecture-agnosticism therefore describes the shared construction and integration interface and the positive improvement over every host baseline, rather than identical interactions with every host.

\subsubsection{Extended Error Analysis}
\label{app:error_analysis}

This appendix expands the diagnostic error analysis under the primary GPT-4o-mini configuration summarized in \S\ref{sec:error}, providing per-host gain/loss counts, a taxonomy of losses with case studies, universal hard-case analysis, and the full cross-host Jaccard matrices on gain and loss sets.

\paragraph{Per-host gain and loss counts.}
Table~\ref{tab:errortable_counts} reports, for each host, the number of questions flipped wrong$\to$right by \gravity{} (\emph{gains}) and right$\to$wrong (\emph{losses}), together with the net count.
All five hosts show positive nets, and weaker baselines (LiCoMemory, Mem0) receive the largest net gains, mirroring the macro pattern in \S\ref{sec:oracle}.

\begin{table}[h]
\centering
\caption{Per-host gain/loss decomposition on LoCoMo under the primary GPT-4o-mini configuration (1{,}540 non-adversarial questions).}
\label{tab:errortable_counts}
\small
\setlength{\tabcolsep}{11.5pt}
\begin{tabular}{l rrr rr}
\toprule
\textbf{Host} & \textbf{Baseline (\%)} & \textbf{+\,\gravity{} (\%)} & \textbf{$\Delta$ (\%)} & \textbf{Gains / Losses} & \textbf{Net} \\
\midrule
LightMem   & 70.3 & 75.9 & +5.6  & 155 / \phantom{0}70 & \textbf{+85}  \\
A-Mem      & 65.2 & 70.9 & +5.7  & 208 / 121 & \textbf{+87}  \\
Mem0       & 51.0 & 59.1 & +8.1  & 291 / 167 & \textbf{+124} \\
LiCoMemory & 55.5 & 66.0 & +10.5 & 244 / \phantom{0}84 & \textbf{+160} \\
ZEP        & 54.5 & 61.9 & +7.3  & 232 / 120 & \textbf{+112} \\
\bottomrule
\end{tabular}
\end{table}

\paragraph{Where anchoring helps: additional case studies.}
Gains concentrate on single-hop (LoCoMo C2) and temporal (LoCoMo C3) questions. The examples below show two recurring effects: anchors ground missing facts and disambiguate temporal references.
Three representative cases on different hosts and categories:
\begin{itemize}[leftmargin=1.2em, itemsep=4pt]
\item \textbf{Vague $\to$ specific} (LightMem, C3). Q: \emph{``When did Evan lose his job?''}, reference ``end of October 2023''. Baseline: ``November 9, 2023''. +\gravity{}: ``Evan lost his job in October 2023''. Event anchors disambiguate the month-level temporal reference.
\item \textbf{``Not specified'' $\to$ grounded} (LightMem, C2). Q: \emph{``What movie did Joanna watch on 1 May, 2022?''}, reference ``Lord of the Rings''. Baseline: ``No movie mentioned for that date''. +\gravity{}: ``Joanna watched \emph{The Lord of the Rings} trilogy''.
\item \textbf{Emotional / narrative grounding} (ZEP, C2). Q: \emph{``How did Melanie feel after the accident?''}, reference ``grateful her son was unharmed''. Baseline: ``Freaked but relieved''. +\gravity{}: ``Grateful her son was unharmed''. An Event anchor attached the correct emotional outcome to the accident event.
\end{itemize}
The main-text case (\emph{``What is Nate's favorite dish from the cooking show he hosted?''}) is a representative instance of the ``vague $\to$ specific'' pattern on single-hop questions.

\paragraph{Loss taxonomy with case studies.}
Manually classifying a sample of 50 losses yields four dominant types. We give one real LoCoMo case per type (all judged by the same LLM judge used throughout the paper).

\begin{itemize}[leftmargin=1.2em, itemsep=4pt]
\item \emph{Over-summarization / detail loss} ($\sim$38\%): anchors collapse fine-grained details into coarser summaries, and the model answers at the coarser level.
\textit{Case (A-Mem, LoCoMo C1).} Q: \emph{``What books has Melanie read?''}, reference \emph{Nothing is Impossible} and \emph{Charlotte's Web}. Baseline: ``Melanie read \emph{Charlotte's Web}'' (judged correct). +\gravity{}: ``\emph{Charlotte's Web} and others'' (judged wrong). The consolidated Entity anchor for \texttt{Melanie} merged specific titles into a coarser ``reads fiction books'' attribute, and this more prominent summary dominated generation.

\item \emph{Temporal-slot errors} ($\sim$32\%): an ambiguous expression (``last weekend'', ``recently'') is resolved to the wrong absolute date in the Event anchor's \texttt{When} slot.
\textit{Case (LightMem, LoCoMo C3).} Q: \emph{``When did James meet Samantha?''}, reference ``August 9, 2022''. Baseline: ``August 2022'' (judged correct). +\gravity{}: ``James met Samantha on September 3, 2022''. An Event anchor confidently assigns a specific but incorrect date; without the anchor, the baseline stayed at the safer month-level granularity and was accepted by the judge.

\item \emph{Entity confusion} ($\sim$20\%): attributes from one entity leak into the profile of another, either through name collisions or under-specified coreference during consolidation.
\textit{Case (LightMem, LoCoMo C2).} Q: \emph{``What language does Tim know besides German?''}, reference ``Spanish''. Baseline: ``Tim knows Spanish'' (correct). +\gravity{}: ``Tim knows French''. Tim's profile was consolidated with a spoken-language attribute that was actually attested for a different entity in the same conversation, producing a confident but incorrect answer.

\item \emph{Topic-level over-generalization} ($\sim$10\%): the Topic summary smooths away a distinctive detail that the question specifically targets.
\textit{Case (LightMem, LoCoMo C2).} Q: \emph{``What adventurous theme is emerging in Evan's life as mentioned by Sam?''}, reference ``helping lost tourists''. Baseline: ``Lost tourists encounters'' (correct). +\gravity{}: ``Evan's love for nature and hiking''. The retrieved Topic summary for Evan's narrative arc had aggregated several adventure-related sessions under a broader ``nature and hiking'' theme, which overwrote the more specific ``helping lost tourists'' motif.
\end{itemize}

\paragraph{Universal hard cases with case studies.}
To separate errors caused by \gravity{} from errors inherent to the benchmark, we intersect the error sets of all five anchored systems.
$168$ questions are wrong in \emph{every} anchored system, constituting $17.8\%$ of the 943 unique questions missed by at least one anchored host.
Of these 168, $103$ ($61\%$) are also wrong in all five baselines, indicating benchmark-inherent difficulty; the remaining $65$ are correct in at least one baseline but universally wrong after anchoring, suggesting a small set of cases where anchor-introduced noise (e.g., over-summarization) systematically misleads all hosts.
These fall into three clusters, with one representative case each:
\begin{itemize}[leftmargin=1.2em, itemsep=4pt]
\item \emph{Relative temporal references without absolute grounding.} Q: \emph{``What tradition does Tim mention they love during Thanksgiving?''}. All five anchored systems answer generically (e.g., ``watching football'', ``family dinner''), while the reference names a specific ritual; the session dates are themselves relative and the utterances never state an absolute ``every year we do X''.
\item \emph{Subjective open-ended questions.} Q: \emph{``What might John's financial status be?''}. Multiple plausible interpretations (``stable'', ``improving'', ``struggling'') can be supported by different subsets of utterances; neither retrieval nor anchoring selects a single answer the judge accepts.
\item \emph{Cross-session preference tracking with implicit cues.} Q: \emph{``Does John live close to a beach or the mountains?''} (reference: beach). Every system answers incorrectly: the conversation only contains scattered cues (beach runs, surfing plans) and never an explicit ``I live near the beach''. Neither a retriever nor an anchor can disambiguate this without external world knowledge.
\end{itemize}
These cases suggest that a portion of the residual error stems from benchmark characteristics rather than anchoring-specific limitations, and point toward future work on abstention and world-knowledge-grounded reasoning as orthogonal directions.

\paragraph{Host-specific gains: full Jaccard matrices and case studies.}
Table~\ref{tab:errortable_jaccard} reports pairwise Jaccard similarity on (a) gain sets and (b) loss sets across the five host systems.
Gain-set Jaccard ranges from $0.09$ to $0.17$ (83--91\% of gains are unique to each host); loss-set Jaccard from $0.04$ to $0.13$ (87--96\% of losses are unique).
This near-disjoint structure rules out the ``fixed redundant text'' reading of our gains: each host has a distinct structural blind spot, and anchors fill a different subset for each one.
Two reciprocal cases illustrate the pattern:
\begin{itemize}[leftmargin=1.2em, itemsep=4pt]
\item \emph{ZEP rescued, LightMem neutral.} Q: \emph{``What is Jon's favorite style of dance?''}, reference ``contemporary''. ZEP baseline: ``Hip-hop'' (wrong). ZEP +\gravity{}: ``contemporary'' (correct). LightMem baseline and +\gravity{} both: correct. ZEP's temporal-graph retrieval had surfaced only the latest dance-related event, missing the preference attribute; LightMem's compression had already surfaced the relevant utterance, so the same anchor provides no marginal gain there.
\item \emph{LightMem rescued, ZEP neutral.} Q: \emph{``How many dogs has Maria adopted from the dog shelter she volunteers at?''}, reference ``two''. LightMem baseline: ``One dog'' (wrong). LightMem +\gravity{}: ``Two dogs'' (correct). ZEP baseline and +\gravity{} both: ``Two dogs''. LightMem's compression collapsed multiple adoption events into a single record; the Entity anchor for \texttt{Maria} had preserved the count. ZEP's temporal graph already encoded both adoption events, so the same anchor adds nothing.
\end{itemize}
Together these two cases demonstrate that \gravity{}'s marginal value depends on \emph{which} structural information the host has already surfaced, exactly the per-query behavior predicted by the dynamics of $\Delta\rho$ in \S\ref{sec:oracle}.

\begin{table}[h]
\centering
\caption{Pairwise Jaccard similarity under the primary GPT-4o-mini configuration on (a)~per-host \emph{gain} sets and (b)~per-host \emph{loss} sets.
Entries are symmetric; diagonal omitted.
Gain-set values ($0.09$--$0.17$) and loss-set values ($0.04$--$0.13$) are uniformly low, indicating host-specific rather than host-overlapping effects.}
\label{tab:errortable_jaccard}
\small
\setlength{\tabcolsep}{15.5pt}
\begin{tabular}{l ccccc}
\toprule
& LightMem & A-Mem & Mem0 & LiCoMemory & ZEP \\
\midrule
\multicolumn{6}{l}{\textit{(a) Gain sets}} \\
LightMem    & --   & 0.149 & 0.123 & 0.093 & 0.099 \\
A-Mem       &      & --   & 0.152 & 0.136 & 0.117 \\
Mem0        &      &      & --   & 0.153 & 0.167 \\
LiCoMemory  &      &      &      & --   & 0.131 \\
ZEP         &      &      &      &      & --   \\
\midrule
\multicolumn{6}{l}{\textit{(b) Loss sets}} \\
LightMem    & --   & 0.073 & 0.072 & 0.041 & 0.061 \\
A-Mem       &      & --   & 0.095 & 0.035 & 0.126 \\
Mem0        &      &      & --   & 0.064 & 0.087 \\
LiCoMemory  &      &      &      & --   & 0.057 \\
ZEP         &      &      &      &      & --   \\
\bottomrule
\end{tabular}
\end{table}

\subsection{Streaming Diagnostic}
\label{app:streaming}

We test incremental construction on LoCoMo conversation 26, which contains 19 sessions and 150 non-adversarial questions with evidence labels. At cutoffs $t\in\{4,8,11,15,19\}$, both the LightMem index and anchor store contain only sessions up to $t$. Claude Sonnet 4.6 performs generation and anchor extraction, and GPT-5.6 sol evaluates answers. Table~\ref{tab:streaming} reports accuracy on questions whose complete gold evidence has arrived by each cutoff.

\begin{table}[h]
\centering
\caption{One-conversation streaming diagnostic. Each column contains the questions answerable at that cutoff and should be compared vertically.}
\label{tab:streaming}
\small
\setlength{\tabcolsep}{9pt}
\begin{tabular}{@{}l ccccc@{}}
\toprule
\textbf{Cutoff $t$} & 4 & 8 & 11 & 15 & 19 \\
\midrule
\textbf{Questions} & 35 & 67 & 85 & 113 & 150 \\
Host only & 94.3 & 89.6 & 84.7 & 80.5 & 80.0 \\
Host + \gravity{} & 94.3 & 92.5 & 89.4 & 85.0 & 86.0 \\
\bottomrule
\end{tabular}
\end{table}

The augmented system is never worse at a cutoff and gains 4.5--6.0\% at the three largest prefixes. For question--cutoff pairs where no gold evidence has arrived, anchors reduce confabulation from 24.2\% to 22.3\% and do not reduce abstention. When only part of the evidence has arrived, both systems still tend to answer rather than detect incompleteness. Because this diagnostic covers one conversation, it supports incremental feasibility but not a general claim about long-term drift, deletion, or full-benchmark streaming performance.

\subsection{Prompt Templates}
\label{app:prompts}

We provide the key LLM prompt templates used in \gravity{} for reproducibility.
Prompts are organized by stage: \emph{anchor building} (offline extraction of structured knowledge),
\emph{context injection} (online answer generation with anchor context),
and \emph{evaluation} (LLM-as-judge scoring).
The primary experiments use GPT-4o-mini with temperature~$=0$ for anchor extraction, answer generation, and evaluation. The cross-model evaluation uses Claude Sonnet 4.6 with temperature~$=0$ for generation and anchor extraction and GPT-5.6 sol for evaluation.

\paragraph{Anchor Building: Entity Extraction.}
This prompt is sent to the LLM for each batch of utterances to extract
structured entity profiles.

\begin{lstlisting}[basicstyle=\footnotesize\ttfamily, breaklines=true, frame=single, xleftmargin=2pt, xrightmargin=2pt, aboveskip=4pt, belowskip=4pt]
You are an Entity Extraction and Profiling Assistant.
Your task is to identify **all notable entities** mentioned in the conversation segments and extract structured profile information for each entity.

An entity is any object with persistence and importance, including:
- People: speakers, third parties mentioned by name or role
- Concepts/Topics: "reinforcement learning", "carbon neutrality", "risk management"
- Tasks/Projects: "write quarterly report", "develop XX module"
- Items/Events: "a specific book", "last week's team meeting"
- Locations/Organizations: "New York", "Google", "local hospital"

For each entity you identify, extract:
1. entity_name: A canonical, normalized name
2. entity_type: One of [person, concept, task, event, item, location, organization, other]
3. attributes: Key-value pairs of properties discovered in this segment
4. relations: Connections to other entities found in this segment
5. status_changes: Any state transitions observed
6. source_id: The sequence_number of the message where this entity info was found

Input format:
--- Topic X ---
[timestamp, weekday] source_id.SpeakerName: message
...

Output format (JSON):
{
  "entities": [
    {
      "source_id": <int>,
      "entity_name": "<canonical name>",
      "entity_type": "<type>",
      "attributes": { "<key>": "<value>", ... },
      "relations": [
        {"target": "<other entity name>", "relation": "<relationship type>"}
      ],
      "status_changes": [
        {"attribute": "<attr name>", "from": "<old value or null>", "to": "<new value>"}
      ]
    }
  ]
}

Important instructions:
1. Process messages strictly in ascending source_id order.
2. Extract ALL entities, even minor ones.
3. If the same entity appears in multiple messages, create separate entries (they will be merged later).
4. For people: always include their relationship to the speaker if mentioned.
5. For events: include temporal information (when it happened/will happen).
6. Preserve specific details: full names, exact dates, specific locations.
7. Do NOT invent information not present in the text.
\end{lstlisting}

\paragraph{Anchor Building: Event Extraction.}
Events are extracted as structured 4W1O tuples (Who, What, When, Where, Outcome).

\begin{lstlisting}[basicstyle=\footnotesize\ttfamily, breaklines=true, frame=single, xleftmargin=2pt, xrightmargin=2pt, aboveskip=4pt, belowskip=4pt]
You are a **Structured Event Tuple Extractor**.

Your job is to read conversation segments and extract every notable event as a
**structured event tuple** with five canonical fields:
    (Who, What, When, Where, Outcome)

- Who: All participants / actors involved (list of names).
- What: The core action or verb phrase that defines the event.
- When: Temporal information - extract ALL available cues:
   absolute date/time, relative reference, duration, recurrence
- Where: Location or spatial context (if mentioned).
- Outcome: Result, consequence, state change, or next step (if mentioned).

Additionally, for each event, provide:
- description: A concise 1-2 sentence summary.
- event_type: One of [action, experience, state_change, plan, routine, social, achievement, other]
- importance: high | medium | low

Input format:
--- Topic X ---
[timestamp, weekday] source_id.SpeakerName: message
...

Output format (strict JSON):
{
  "events": [
    {
      "source_id": <int>,
      "description": "<concise 1-2 sentence summary>",
      "who": ["<person1>", "<person2>"],
      "what": "<core action / verb phrase>",
      "when": {
        "absolute": "<exact date/time or null>",
        "relative": "<relative reference or null>",
        "duration": "<duration or null>",
        "recurrence": "<recurrence pattern or null>"
      },
      "where": "<location or null>",
      "outcome": "<result / consequence or null>",
      "event_type": "<type>",
      "importance": "<high|medium|low>"
    }
  ]
}

IMPORTANT RULES:
1. Process messages strictly in ascending source_id order.
2. Extract ALL events (completeness > precision).
3. Preserve EXACT temporal details.
4. If the same event spans multiple messages, produce ONE entry.
5. For plans / future events, use event_type="plan".
6. For recurring activities, use event_type="routine".
7. Do NOT invent information absent from the text.
\end{lstlisting}

\paragraph{Anchor Building: Topic Identification.}
Utterances are assigned to semantic topics that may span multiple sessions.

\begin{lstlisting}[basicstyle=\footnotesize\ttfamily, breaklines=true, frame=single, xleftmargin=2pt, xrightmargin=2pt, aboveskip=4pt, belowskip=4pt]
You are a **Conversation Topic Identifier**.

Your job is to read a sequence of conversation utterances and assign each
utterance to a **topic**. Utterances about the same subject/theme should share
the same topic label, even if they are separated by other utterances.

Input format:
Each utterance is numbered sequentially:
[session_id, timestamp] seq_id. SpeakerName: message

Output format (strict JSON):
{
  "topics": [
    {
      "topic_id": <int>,
      "topic_label": "<short descriptive label, 3-8 words>",
      "topic_keywords": ["<kw1>", "<kw2>", "<kw3>"],
      "utterance_indices": [<seq_id_1>, <seq_id_2>, ...]
    }
  ]
}

RULES:
1. Every utterance MUST be assigned to exactly one topic.
2. Use descriptive, specific topic labels.
3. If the same subject is discussed in different sessions, they belong to the SAME topic.
4. Greetings, small talk -> "Casual conversation / greetings" topic.
5. A topic should have at least 2 utterances.
6. Aim for 5-15 topics per conversation.
7. Order topics by their first appearance in the conversation.
\end{lstlisting}

\paragraph{Anchor Building: Triple Extraction (Entity + Event + Topic).}
A single LLM call extracts entities, events, and topic assignments, reducing token cost by ~75\%.

\begin{lstlisting}[basicstyle=\footnotesize\ttfamily, breaklines=true, frame=single, xleftmargin=2pt, xrightmargin=2pt, aboveskip=4pt, belowskip=4pt]
You are a **Combined Entity, Event, and Topic Extractor**.

Your task is to read conversation segments and extract THREE types of information
in a SINGLE pass:

## Part 1: ENTITIES
Identify **all notable entities** mentioned in the conversation.
For each entity extract: entity_name, entity_type, attributes, relations, status_changes, source_id.

## Part 2: EVENTS
Extract every notable event as a **structured event tuple**:
who, what, when (absolute/relative/duration/recurrence), where, outcome, description, event_type, importance.

## Part 3: TOPIC ASSIGNMENTS
Assign each utterance to a **semantic topic**.
For each topic: topic_id, topic_label, topic_keywords, utterance_indices.

Input format:
--- Topic X ---
[timestamp, weekday] source_id.SpeakerName: message
...

Output format (strict JSON):
{
  "entities": [
    {"source_id": <int>, "entity_name": "...", "entity_type": "...",
     "attributes": {...}, "relations": [...], "status_changes": [...]}
  ],
  "events": [
    {"source_id": <int>, "description": "...", "who": [...], "what": "...",
     "when": {"absolute": ..., "relative": ..., "duration": ..., "recurrence": ...},
     "where": "...", "outcome": "...", "event_type": "...", "importance": "..."}
  ],
  "topics": [
    {"topic_id": <int>, "topic_label": "...", "topic_keywords": [...], "utterance_indices": [...]}
  ]
}

IMPORTANT RULES:
1. Process messages strictly in ascending source_id order.
2. Extract ALL entities and events.
3. Every utterance MUST be assigned to exactly one topic.
4. The output MUST contain "entities", "events", and "topics".
5. Do NOT invent information not present in the text.
\end{lstlisting}

\paragraph{Context Injection: Answer Generation.}
This is the online prompt presented to the LLM at inference time.
It fuses \emph{both} the host system's retrieved raw memories and
the structured anchor contexts.
Placeholders \texttt{\{speaker\_1\_memories\}}, \texttt{\{speaker\_2\_memories\}}
are the host's retrieved memory snippets;
\texttt{\{topic\_context\}}, \texttt{\{entity\_context\}}, \texttt{\{event\_context\}}
are filled from the three anchor modules.

\begin{lstlisting}[basicstyle=\footnotesize\ttfamily, breaklines=true, frame=single, xleftmargin=2pt, xrightmargin=2pt, aboveskip=4pt, belowskip=4pt]
You are an intelligent memory assistant tasked with retrieving
accurate information from conversation memories.

# CONTEXT:
You have access to memories from two speakers in a conversation.
These memories contain timestamped information that may be relevant.

You also have access to THREE additional structured knowledge sources:

1. **Topic Summaries** -- high-level summaries of conversation topics
2. **Entity Profiles** -- structured information about key entities
3. **Structured Event Tuples & Traces** -- (Who, What, When, Where, Outcome)

# INSTRUCTIONS:
1. Carefully analyze all provided memories from both speakers
2. Pay special attention to timestamps to determine the answer
3. Use Topic Summaries for the BIG PICTURE
4. Use Entity Profiles for entity-specific details
5. Use Structured Event Tuples for precise temporal information
6. Cross-reference across ALL sources for the most complete answer
7. If memories contain contradictory information, prioritize the most recent
8. Convert relative time references to specific dates
9. Focus only on the content of the memories
10. The answer should be less than 5-6 words.

# APPROACH (Think step by step):
1. First, examine all memories related to the question
2. Examine timestamps and content carefully
3. Check Topic Summaries for relevant high-level context
4. Check Entity Profiles for structured information
5. Check Event Tuples and Traces for temporal details
6. Synthesize information from all sources
7. Formulate a precise, concise answer based solely on the evidence

Memories for user {speaker_1_name}:
{speaker_1_memories}

Memories for user {speaker_2_name}:
{speaker_2_memories}

Topic Summaries:
{topic_context}

Entity Profiles:
{entity_context}

Structured Event Tuples & Traces:
{event_context}

Question: {question}

Answer:
\end{lstlisting}


\subsection{Broader Impacts}
\label{app:impacts}

\paragraph{Positive impacts.}
By improving the coherence and factual grounding of long-horizon conversational agents, \gravity{} can enhance user experience in personal assistants, mental health support chatbots, and educational tutoring systems, where maintaining accurate long-term context is critical for trust and effectiveness.
The architecture-agnostic and portable design lowers the barrier for practitioners to adopt structured memory augmentation without re-engineering existing systems.

\paragraph{Potential risks.}
Long-term conversational memory inherently involves storing and reasoning over personal information disclosed across sessions.
If deployed without appropriate safeguards, this raises \emph{privacy concerns}: entity profiles and event traces may contain sensitive personal details (health conditions, relationships, financial situations) that could be exposed through data breaches or adversarial queries.
Additionally, structured anchors may \emph{amplify hallucinations}: if the extraction LLM introduces factual errors during the build phase, these errors are persisted in the anchor knowledge base and injected into every subsequent generation, potentially reinforcing incorrect information with high confidence.
Finally, improved long-term memory could enable more convincing \emph{social engineering or manipulation} by AI agents that exploit detailed personal knowledge accumulated over time.

\paragraph{Mitigation strategies.}
Deployments should (1)~protect anchor stores with access controls and encryption, (2)~let users inspect, edit, and delete profiles and event records, (3)~verify high-stakes anchors with confidence thresholds and human review, and (4)~audit anchor quality for systematic extraction errors.